\documentclass[12pt]{article}

\usepackage{latexsym,epsfig,graphicx,amsmath,amssymb,amscd,multirow,bbm,chicago}
\usepackage{chicago,psfrag,paralist,dsfont,url,comment}
\usepackage{subfig}
\usepackage{graphicx}
\usepackage[titletoc]{appendix}

\usepackage{epstopdf}

\usepackage[multidot]{grffile}

\renewcommand{\min}{\textrm{min}}
\renewcommand{\exp}{\textrm{exp}}
\newcommand{\bx}{\boldsymbol{x}}
\newcommand{\bt}{\boldsymbol{t}}
\newcommand{\bh}{\boldsymbol{h}}

\newcommand{\bo}{\boldsymbol{o}}

\newcommand{\SEV}{\textrm{SEV}}

\newcommand{\bw}{\boldsymbol{w}}

\newcommand{\bbeta}{\boldsymbol{\beta}}

\newcommand{\MVN}{\textrm{MVN}}

\begin{document}

\title{SSH-Net: A Deep Neural Network for Predicting Failure Time Distribution Functions under Competing Risks with Application to GPU Data}

\author{
Jie Min$^{1}$, Yueyao Wang$^{2}$\footnote{Corresponding Author. Email: yueyaowang@mail.zjgsu.edu.cn}, and Mengkun Chen$^{3}$\\[1.5ex]
	{\small $^1$Department of Mathematics \& Statistics, University of South Florida, Tampa, FL}\\[0.05ex]
	{\small $^2$School of Statistics and Data Science, Zhejiang Gongshang University, Hangzhou, China}\\[0.05ex]
	{\small $^{3}$Department of Statistics, Virginia Tech, Blacksburg, VA}\\[0.05ex]
}

\date{}
\maketitle

\begin{abstract}
Competing risks are commonly observed in engineering fields, and can bring challenges to time-to-event data modeling when the application scenarios are complicated. Recently, deep neural networks have received great attention for prediction with competing risks, due to their flexibility and high learning capability. However, the complexity of neural network structure brings extra difficulty in hyperparameter tuning based on different data inputs. Additionally, when an engineered system has complex physical structures with multiple hierarchical levels, treating all structural levels as a single group of inputs may fail to capture critical information. To address the issues, we propose a Structured Segmented Hazard Deep Neural Network (SSH-Net) for failure time prediction under cause-specific competing risks framework. Our approach associates neural network structure with data structures, and allows different covariate groups to impact the failure prediction through separate sub-networks. The neural network is constructed based on a cause-specific competing risks model. The SSH-Net outputs cause-specific hazard functions, and utilizes the penalized log-likelihood as the loss function. The prediction accuracy of SSH-Net is validated through simulation studies by evaluating the Brier score, the area under receiver operating characteristic curves (AUC), and the root mean square error (RMSE) of the predicted cause-specific cumulative incident function. We further demonstrate the model's ability to predict failure time distribution functions using the Titan GPU failure time data.


\end{abstract}

\section{Introduction}\label{sec:introduction}
 \subsection{Motivation}
Failure time modeling and prediction for time-to-event data are commonly seen in medical research including cancer studies (\shortciteNP{shahraki2023time}) and analysis of Alzheimer's progression (\shortciteNP{sharma2021time}), as well as the reliability analysis of engineering systems and their component (\shortciteNP{man2018remaining}).  In such studies, more than one causes of failures may exist, where observing the failure time from one cause prevents observing the failure from other causes, leading to the setting of competing risks. With competing risks, distribution functions of the failure times for different failure types can be used to provide probabilistic predictions.  In modeling time-to-event data with competing risks, classical survival models typically rely on parametric assumptions regarding the distribution of failure times or certain structure of covariate effects. During recent years, to relax such parametric assumptions and to leverage the prediction power of deep learning models, several neural networks are proposed for prediction with competing risks. One drawback of using neural networks is the prediction performance can be sensitive to the number of nodes and layers in a general proposed neural network structure, given different data. Selecting the number of nodes and layers are commonly included in hyperparameter tuning based on grid search. When the neural network structure cannot be clearly explained, the searching grid is determined based on past experience, leading to the risk that the best set of nodes and layers are not included in the grid. Additionally, when covariates contain special features, such as a hierarchical structure based on the system, it can be difficult for a general neural network to capture these information. To solve these issues, we propose a Structured Segmented Hazard Deep Neural Network (SSH-Net) for predicting the distribution functions of failure times under cause-specific competing risks framework, in which part of the neural network structure is associated with the data structure. The SSH-Net structure has its explanation based on the data, which can be used in hyperparameter tuning. In addition, covariates with different features are processed separately in SSH-Net with different sub-networks, which can improve the prediction performance. A detailed introduction of the neural network structure is presented in Section~\ref{sec:structure}.

We illustrate the usage of SSH-Net using the failure time prediction of graphics process units (GPUs) inside supercomputers. As GPUs within supercomputers are intensively used in training models with large data, the reliability of GPUs is of interest during recent years. GPUs in supercomputers can fail due to multiple types of errors, including connection error and memory error, which implies the presence of competing risks. Additionally, the failure times of GPUs can be influenced by the locations of cabinets and the positions of GPUs inside cabinets (\shortciteNP{ostrouchov2020gpu}), resulting in covariates with hierarchical levels and spatial location information that require extra attention in prediction. One popular dataset used in GPU reliability analysis is the GPU failure time data in the Cray XK7 Titan supercomputer. The performance of SSH-Net is demonstrated on the Titan GPU data.

\subsection{Related Work}
It is well known that neural networks have the ability to learn non-linear relationships between covariates and response without parametric assumptions. To leverage this advantage, multiple deep neural networks have been developed to model time-to-event data.  For example, Cox regression without parametric assumptions is extended through neural networks (\shortciteNP{katzman2018deepsurv} and \shortciteNP{kvamme2019time}). Recurrent neural networks are proposed to learn the probability mass function (pmf) of discretized survival time (\shortciteNP{giunchiglia2018rnn}) or the hazard rate of discretized survival time (\shortciteNP{ren2019deep}). Additionally, neural networks using piecewise constant hazard functions as outputs are developed in \shortciteN{fornili2014piecewise} and  \citeN{kvamme2021continuous} to avoid discretizing continuous time.

While the above mentioned methods can handle only one type of failure, deep learning methods are also explored in estimating the distribution functions with competing risks. Among the rich literature, one of the popular neural networks for competing risks is DeepHit, proposed by \shortciteN{lee2018deephit}. The model treats the failure times as discrete random variables and uses log-likelihood loss together with a penalization on incorrect ordering pairs to train the neural network in learning the cause-specific pmf of different failure modes. Also treating failure times as discrete random variables, \shortciteN{gupta2019cresa} combined Long Short-Term Memory (LSTM) layers with fully connected layers to improve the prediction. On the other hand, by adding extra ordinary differential equation (ODE) solver layers onto a similar neural network structure proposed by \shortciteN{lee2018deephit}, \shortciteN{huang2021deepcompete} treated the failure times as continuous random variable, and used log-likelihood loss in training. Also using ODEs and log-likelihood loss,  \citeN{danks2022derivative} proposed to output the cause-specific cumulative incident functions (CIFs), together with the failure probability for different failure types.  Additionally, \shortciteN{wang2022survtrace}  modified transformers to learn the cause-specific hazard functions with piecewise constant assumption and applied inverse propensity score to deal with extremely imbalanced censor rate across multiple failure types, with a loss function constructed based on multi-task learning. More recently, \shortciteN{jeanselme2023neural} proposed the Neural Fine Gray (NFG), which is a computationally efficient monotonic neural network to predict the CIFs. Other than the neural networks that output values of distribution functions, \shortciteN{rahman2021deeppseudo} proposed to use pseudo values calculated based on the Aalen-Johansen estimators as neural network outputs. Additionally, \shortciteN{lee2019dynamic} extended the neural network structure in \shortciteN{lee2018deephit} to account for longitudinal covariates. \shortciteN{hong2022deep} utilized contrastive learning in prediction with longitudinal covariates.

Compared to the existing neural network models, our proposed SSH-Net has a more interpretable neural network structure associated with the data structure, which can be used to assist hyperparameter tuning. The SSH-Net avoids discretizing continuous time by using piecewise-constant hazard functions. Furthermore, a smoothness penalty term is added to the loss function to avoid overfitting, which was not considered in previous literature using piecewise-constant hazard functions. The high prediction accuracy for SSH-Net is shown using criteria including the Brier score, the area under receiver operating characteristic curves (AUC), and root mean squared errors (RMSE) of CIFs, based on the simulation data and the Titan GPU data introduced in Section~\ref{sec:data}. 

On the application side, the reliability of GPUs in supercomputers has received increasing attention in recent years due to the importance of GPUs in training models on large-scale datasets. In the current literature, the performance of neural network models are commonly compared based on cancer research data such as the Surveillance, Epidemiology, and End Results Program (SEER) dataset published by \citeNP{seer}\unskip. In this paper, the performance of neural network prediction is first evaluated based on a GPU failure time data, which have a special covariate structure, including spatial location information and multiple hierarchical levels. In analyzing the reliability of GPUs inside supercomputers, \shortciteN{Min03072025} developed an accelerated failure time model with spatially correlated random effects and mixture distribution, assuming Weibull or Lognormal distributions in the failure time modeling. \shortciteN{clark2025modeling} further associated multiple types of distance functions with the accelerated failure time model. \shortciteN{wang2025use} investigated the usage of computationally efficient variational inference method in the GPU lifetime analysis. Additionally, \shortciteN{nie2018machine} evaluated performance of machine learning models such as support vector machine and gradient boosting decision tree to predict the single-bit errors, without competing risks.  Other statistical models related to time-to-event analysis with competing risks and spatial location information including \shortciteN{hesam2018cause} and \shortciteN{momenyan2020modeling}, where the location information is treated as random effects and is added to the hazard functions.

\subsection{Contributions}
The main contributions of this paper are as follows. First, we propose SSH-Net, a novel neural network structure that can accommodate complex data structures, such as hierarchical covariates structure, and associates the network structure with underlying data structure, thereby easing hyperparameter tuning. Also, we implement a log-likelihood loss with a smooth penalty to handle continuous failure time and reduce overfitting. Additionally, we design simulation studies that can provide fair comparison among several candidate neural networks without favoring their assumptions. The simulation study design also enables the usage of RMSE to better evaluate the performance of the neural networks, while existing literature focuses on time-dependent Brier score and the weighted time-dependent AUC. The simulation study results illustrate the advantage of SSH-Net compared to candidate models. Lastly, we apply the SSH-Net to analyze the Titan GPU data, showing the prediction of CIFs, the important factors that influence the GPU failure times, and the relative risk among multiple GPU failure types.



\subsection{Overview}
The rest of the paper is arranged as follows. Section~\ref{sec:model} presents the details of the proposed SSH-Net model, including the data notation and cause-specific model, and the SSH-Net network structure, loss function, and hyperparameter selection. Section~\ref{sec:sim} demonstrates the performance of SSH-Net using simulation studies. Section~\ref{sec:real} applies the SSH-Net to analyze the Titan GPU data. Section~\ref{sec:conclusion} summarizes the conclusions and future research areas.


\section{Methodology}\label{sec:model}

\subsection{Preliminaries}

In a time-to-event data analysis, for each unit $i$, let $T_{i}$ denote the continuous failure time, $C_i$ denote the censoring time, and $t_i$ be the observed value of $\widetilde T_i = \min(T_i, C_i)$, which is the observed failure time or censoring time. Let $\bx_i = (\bx_{ic}', \bx_{il}')' =  (x_{ic_1},\ldots, x_{ic_{p_c}}, x_{il_1},\ldots, x_{il_{p_l}})'$ be the covariates corresponding to unit $i$. In particular, let $\bx_{il}$ denote the set of ``global'' covariates and $\bx_{ic}$ denote the covariates at a lower hierarchical level in a system with a hierarchical structure. The $p_l$ and $p_c$ are the dimensions of $\bx_{il}$ and $\bx_{ic}$, respectively. Let $G_i$  denote the failure type of unit $i$, where $G_i = 1, \ldots, K$, and $K$ is the total number of types of failures. Let $\delta_{ik}$ denote the event indicator. That is, $\delta_{ik}=1$ if unit $i$ fails because of failure mode $k$, and $0$ otherwise. Note that for a censored unit, $\delta_{ik} = 0$ for all $k = 1 ,\ldots, K$. In summary, we denote the time-to-event data as $\bt = \{ t_i, \bx_i, \delta_{ik}\}$, i = $1, 2, \ldots, n$, where $n$ is the total number of units, and $k = 1, 2, \ldots, K$.

In a cause-specific competing risks model, the overall hazard function for unit $i$ is defined as 
\begin{align*}
	h_i(t ) = \ \lim_{\delta \to 0} \frac{\Pr(T_i \leq t+ \delta \mid T_i> t)}{\delta},
\end{align*}
and the cause-specific hazard function for failure mode $k$ is
\begin{align*}
	h_i(t,k ) = \ \lim_{\delta \to 0} \frac{\Pr(G_i = k, T_i \leq t+ \delta \mid T_i> t)}{\delta}.
\end{align*}
In particular, for a continuous failure time, the cause-specific hazard can be expressed as
\begin{align}
	h_i(t,k ) = \frac{f_i(t,k)}{S_i(t)}, \label{formula: haz1}
\end{align}
where $f_i(t,k)$ is the cause-specific probability density function (pdf) such that $$f_i(t,k) = \frac{dF_i(t,k)}{dt},$$ and $F_i(t,k)$ is the CIF such that $F_i(t,k) = \Pr(G_i=k, T_i\leq t)$.  The $S_i(t)$ in \eqref{formula: haz1} is the overall survival function, and $S_i(t) = 1 - F_i(t)$, where $F_i(t) = \Pr(T_i \leq t)$ is the overall cumulative distribution function (cdf). In particular, the relationship between cause-sepcific distribution functions and overall distribution functions are 
\begin{align*}
	h_i(t) = \sum_{k = 1}^K h_i(t,k) \;,  F_i(t) = \sum_{k = 1}^K F_i(t,k ), \; \textrm{and} \;  S_i(t) = \sum_{k = 1}^K S_i(t,k ).
\end{align*} 
Note that the failure probability for failure mode $k$ is $ P(G_i=k)$, and $S_i(t,k) = P(G_i = k) - F_i(t,k)$. Also, the overall survival can be expressed as
\begin{align*}
	S_i(t) = \sum_{k = 1}^K S_i(t,k ) =  \sum_{k = 1}^K \textrm{exp}\left[- \int_{0}^t h_i(u, k) du\right].
\end{align*}
The loss function for the proposed SSH-Net is constructed based on the cause-sepcific hazard and the overall survival function, and will be discussed in detail in Section~\ref{sec:loss}.

\subsection{The proposed Neural Network Structure}\label{sec:structure}
\begin{figure}
	\centering
	\includegraphics[width = 0.8\textwidth]{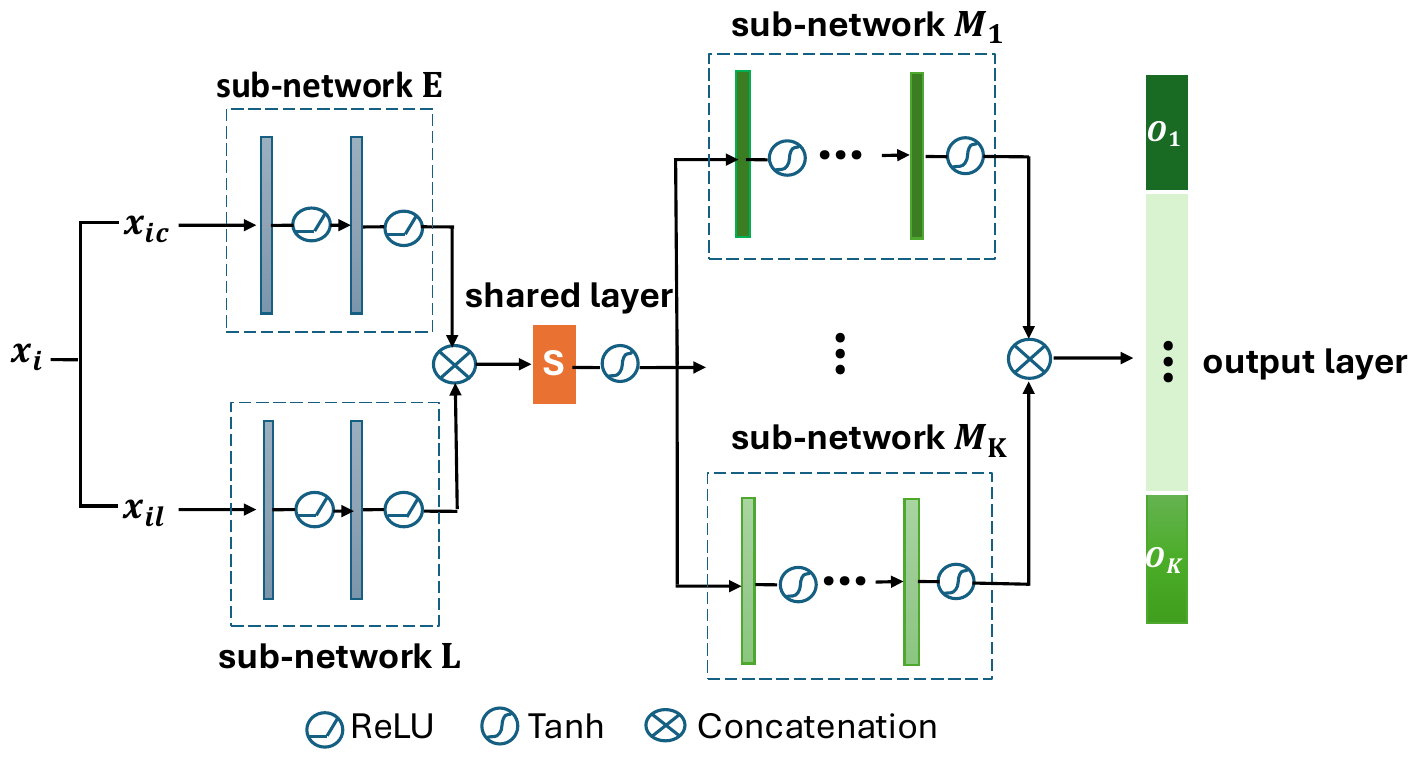}
	\caption{Visualization of the neural network structure of SSH-Net, where E represents the sub-network for $\bx_{ic}$, L represents the sub-network for $\bx_{il}$, S represents the shared layer, $M_1,\ldots,M_K$ represent the sub-networks for multiple failure modes, and $O_1,\ldots, O_K$ represent output layers, which are the constant log cause-specific hazards.}\label{fig:structure}
\end{figure}

The structure of the proposed SSH-Net is shown in Figure~\ref{fig:structure}.   Specifically, let
\begin{align}
	h(t,k\mid\bx_i) =h_i(t, k ) =  \lim_{\delta \to 0} \frac{\Pr(G_i = k, T \leq t+ \delta  \mid T > t, \bx_i)}{\delta} \label{formula:haz}
\end{align}
be the cause-specific hazard function for failure mode $k$ of unit $i$ with covariates $\bx_i$. We assume that $h(t, k\mid \bx_i)$ is piecewise constant on the time interval $(0, \tau)$, where $\tau$ is the length of the study time. For example, $\tau$ can be the maximum observed failure or censor time in the training dataset, or a time that is slightly longer than the maximum observed time. Equally spaced time bins are then built based on the time interval $(0,\tau)$, which are, 
$$(0, \Delta_k), (\Delta_k, 2\Delta_k), \ldots, ((J_k-1)\Delta_k, \tau),$$
where $\Delta_k$ is the length of the bins for failure type $k$, and $J_k$ is the total number of bins for failure type $k$. Note that the different number of bins across failure types can lead to extra flexibility in estimation, and can reduce the influence from imbalanced censor rates across the failure types. Although the best $J_k$ needs to be chosen using hyperparameter selection, the range of the $J_k$ grid can be determined based on common knowledge related to the study, as discussed in Section~\ref{sec:hyper}. With the piecewise constant assumption, the segmented cause-specific hazard function in \eqref{formula:haz} can be written as 
\begin{align}
	h(t,k\mid \bx_i) = \sum_{j = 1}^{J_k} \mathbbm{1} { \left\{ (j-1)\Delta_k< t \leq j\Delta_{k} \right\}}h_{jk\mid \bx_i}, \quad j = 1, 2, \ldots, J_k, \label{formula:h2}
\end{align}
where $h_{jk \mid \bx_i}$ is the constant hazard. We let the logarithm of $h_{jk \mid \bx_i}$ be the outputs of the neural network. 

To deal with special covariate features, different sub-networks are built for the ``global'' covariate $\bx_{il}$ and the lower hierarchical level covariate $\bx_{ic}$. In particular, we let $L$ represent the sub-network for $\bx_{il}$, and  $E$ represents the sub-network for $\bx_{ic}$. In $L$ and $E$, the number of nodes are associated with $p_l$ and $p_c$, which are the dimensions of $\bx_{il}$ and $\bx_{ic}$, respectively. In each sub-network, two layers with the same number of nodes are used to learn both the main effects and the second order interactions among the covariates using fully connected layers. Higher order interactions are not considered as they have weaker impact on the failure times. A common shared layer $S$ is then added on top of $E$ and $L$. We propose to use relatively small number of nodes in $S$ compared to $p_{c} + p_{l}$, so that $S$ serves as a method for ``data reduction'', and is used to extract useful information from $E$ and $L$. 


The information obtained by $S$ is passed to the fully connected sub-networks $M_1,\ldots, M_K$, which are sub-networks for the $K$ failure types. In a sub-network $M_k$, the nodes can be viewed as latent variables that influence the cause-specific hazard functions in \eqref{formula:haz}. With the piecewise constant assumption, we expect the number of latent variables stay in the same scope as the number of bins $J_k$. If there are many latent variables, it is less likely that the cause-specific hazard remains constant within a bin. On the other hand, if there are few latent variables, a smaller number of bins should be used. Therefore, we propose to use a value close to the median of the grid for $J_k$ as the number of nodes in $M_k$. 


The activation functions in $E$ and $L$ are chosen to be ReLU to avoid vanishing gradient problem, and the activation functions in $S$ and $M_1, \ldots, M_K$ are Tanh, which can potentially reduce the bias in neural network prediction as the output of Tanh is symmetric about zero (\shortciteNP{li2020combine}).


\subsection{Loss Function of the SSH-Net}\label{sec:loss}

To avoid wiggly behavior of the cause-specific hazard functions, we use the minus penalized log likelihood as the loss function, where the penalty is added to the difference of adjacent constant hazards $h_{jk \mid\bx_i}$. This method is also used with P-splines, as illustrated in \citeN{marra2020copula}. Let $\bo_{ik} = (o_{1k \mid\bx_i},\ldots, o_{J_k k \mid\bx_i})'$, where $\bo_{ik}$ is the SSH-Net output and $o_{jk \mid\bx_i} = \log(h_{jk \mid\bx_i})$, and let $\bh_{ik} = ({h_{1k \mid\bx_i}},\ldots, h_{J_k k \mid\bx_i})'$.
The minus penalized log likelihood loss is calculated as
\begin{align}
	l = - \sum_{k = 1}^K \sum_{i = 1}^n \left[ \delta_{ik} \log [h(t_i,k \mid\bx_i) ]-  H(t_i,k\mid\bx_i) - \lambda_k {\bh_{ik} }' D {\bh_{ik}}\right],  \label{formula:lik}
\end{align}
where $H(t,k \mid \bx_i) $ is the cause-specific cumulative hazard function such that 
$$H(t,k \mid \bx_i)  = \int_{0}^t h(u,k\mid\bx_i) du.$$ 
Based on the expression of \eqref{formula:h2} for $h(t_i, k \mid \bx_i)$, 
\begin{align*}
&	H(t,k \mid \bx_i) =\sum_{j = 1}^{ \lfloor t/\Delta_k \rfloor} \Delta_k h_{jk\mid \bx_i} + \left(t -  \lfloor t/\Delta_k \rfloor \Delta_k  \right)h_{  \lfloor t/\Delta_k \rfloor  k\mid \bx_i}, \quad t> \Delta_k,\\
&	\textrm{and} \quad 	H(t,k \mid \bx_i) = th_{ 1 k\mid \bx_i} \quad 0 < t \leq \Delta_k,
\end{align*} 
where $ \lfloor t/\Delta_k \rfloor$ denotes the floor of $t/\Delta_k$. The $D$ matrix in \eqref{formula:lik} is a $J_k \times J_k$ matrix used to penalize the differences among adjacent constant hazard rates. In particular,
\begin{align*}
	D = 
	\begin{pmatrix}
	1& -1 & 0 & \ldots & 0 & 0\\
	-1 & 2 & -1 & \ldots & 0 & 0 \\
	\vdots & \vdots &\vdots & & \vdots & \vdots\\
	0 & 0 & 0 & \ldots & 1 & -1
	\end{pmatrix}.
\end{align*}
Additionally, $\lambda_k$ is the hyperparameter related to the smooth penalty of failure type $k$.

\subsection{Hyperparameter Selection}\label{sec:hyper}
The three important sets of hyperparameters in the proposed neural network model are the number of bins $J_k$, the smooth penalty $\lambda_k$, and the number of layers in each sub-network $M_k$.  The hyperparameter selection can be performed using penalized log likelihood on the validation set based on grid search. In selecting the $\lambda_k$, we use $\lambda_k = \lambda^*_k \times (\sum_{i = 1}^n \delta_{ik})^{1/3}$, as the hyperparameter related to the penalty should increase with the number of observations in the dataset. Since failed units include more information related to the censored units, we use the number of failed units when adjusting the $\lambda_k$. In this paper, the grid for $\lambda^*_k$ is $\{0, 0.5, 1, 1.5 \} $. For the number of layers of the sub-networks for failure modes, we consider a grid of $\{1, 2, 3\}$. 

Additionally, as the structure of SSH-Net associates with the data structure, the choice of grid of $J_k$ can be guided based on the underlying system and the study period $\tau$, which eases the process of hyperparameter tuning. Section~\ref{sec:dataap} illustrates the details of hyperparameter tuning of $J_k$ using the Titan GPU data as an example.
%


\section{ Simulation Studies}\label{sec:sim}	
We use simulation to illustrate the performance of SSH-Net and show its advantage related to candidate neural networks models. In particular, although we delay the introduction of the Titan GPU data to Section~\ref{sec:data}, we generate simulated data that are similar to the Titan GPU data to provide a realistic simulation setting. Note that the design of simulation provides fair comparison among candidate models and does not favor the SSH-Net, since the data is generated based on a latent failure time model introduced in Appendix~\ref{apd1}, rather than a cause-specific model. The true distribution functions from the latent failure time model can be transformed to match the outputs of SSH-Net and other neural network models, as shown in Appendix~\ref{apd2}. Multiple simulation scenarios with different number of units and different censor rates are considered to explore the influence of the number of units and censor rates on the prediction performance. Under each scenario, $100$ datasets are generated. A detailed introduction about the simulation setting is presented in Section~\ref{sec:setting}.


We compare the performance of the proposed SSH-Net with the popular neural network DeepHit (\shortciteNP{lee2018deephit}) and the recently proposed NFG model (\shortciteNP{jeanselme2023neural}). Hyperparameter selection is performed based on the validation set for each of the neural network.
For DeepHit, the number of layers for cause-specific and shared sub-networks are chosen from $\{1,2,3,5\}$, number of nodes is selected from $\{50, 100, 200, 300\}$, and the penalty for ranking loss is selected from $\{0, 0.01, 0.05\}$. For NFG, the number of hidden layers for the embedding network, sub-distribution network, and balancing network are chosen from $\{1,2,3,4\}$,  and the number of nodes are chosen from $\{50, 100\}$.  For SSH-Net, the number of bins is selected from $\{10,20,30,50\}$, the number of layers in $M_k$ is selected from $\{1, 2, 3\}$, and the number of nodes in $M_k$ and the output layer is fixed at $50$ based on the structure of the dataset. A detailed explanation of the chosen grids for SSH-Net is deferred to Section~\ref{sec:dataap}, which presents the real GPU reliability application, as the simulated data are generated based on the Titan GPU data. Additionally, the smooth penalty is selected from $\{0, 0.5, 1, 1.5\}$. Table~\ref{tab:trainset} shows the optimization settings in training all the neural networks.  

 To evaluate the performance of the neural networks, we use three criteria, which are the RMSE of CIFs, the time-dependent weighted Brier score, and the weighted time-dependent AUC that is used to measure the concordance of the ranking between predicted failure times and the observed times (\shortciteNP{antolini2005time}).  The calculation of evaluation criteria is discussed in Section~\ref{sec:eva}.

\subsection{Simulation Settings}\label{sec:setting}

In the simulation, we study the influence of two factors on the neural network performance, which are the number of units $n$, and the censor rates of the two types of failures.  In particular, $8$ scenarios are considered, and Table~\ref{tab:simu.setting} shows the number of units and the average censor rates of the $100$ datasets for each simulation scenario. 

To generate datasets similar to the Titan GPU data while maintaining a fair comparison among candidate models, we employ the latent failure time model proposed by \shortciteN{Min03072025}, which does not inherently favor the cause-specific framework used in SSH-Net. A brief introduction about the latent failure time model, the true parameter setup for the simulation, and the relationship between the latent failure time model and the cause-specific model is presented in the Appendix~\ref{apd1}. In simulation, the covariates $\bx_i$ and insert dates of GPUs are sampled directly from the Titan GPU dataset. The censoring rate is controlled by varying the shutdown date of the supercomputer, where a later shutdown date results in a lower censoring rate, as shown in Table~\ref{tab:simu.setting}.

In each simulation run, the dataset is divided into training set, validation set, and test set, with proportions 64\%, 16\%, and 20\%, respectively, stratified by the failure type labels, which are censored, Failure Type 1, and Failure Type 2.

\begin{table}[]
	\centering
	\caption{Number of GPU units and average censor rates for the two failure types in different simulation scenarios.}\label{tab:simu.setting}
	\begin{tabular}{c|cccc}
		\hline\hline
		Shut Down Date  & 2020-01-01 &  2019-01-01 &  2018-01-01 & 2017-01-01 \\
		\hline
		\multicolumn{5}{c}{$n$ = 10,000}\\
		\hline
		Failure Type 1 Censor Rate & 84.02 \%       & 89.32 \%    & 93.85\%    & 97.42\%    \\
		Failure Type 2 Censor Rate & 62.37 \%       & 70.53\%   & 81.49 \%    & 92.35 \%  \\
		\hline
		\multicolumn{5}{c}{$n$ = 5,000}\\
		\hline
		Failure Type 1 Censor Rate & 81.50 \%       & 87.07 \%    & 92.32\%    & 96.49\%    \\
		Failure Type 2 Censor Rate & 62.06\%       & 70.24\%   & 80.85 \%    & 91.83\%  \\
		\hline\hline
	\end{tabular}
\end{table}

\begin{table}
	\caption{Optimization setting for SSH-Net, NFG, and DeepHit in training.}\label{tab:trainset}
	\centering
	\begin{tabular}{c|ccccc}
		\hline\hline
		Neural Network & Batch Size & Optimizer & Learning Rate &  Epochs & Early Stop\\ \hline
		SSH-Net & 100 & AdamW & $1e-3$ & 100& 10\\
		NFG & 100 &   Adam & $\{1e-3, 1e-4\}$ & 200 & 3 \\
		DeepHit & 100 &  Adam & $1e-4$& 500 & 5 \\ \hline\hline
	\end{tabular}
\end{table}

\subsection{Evaluation Criteria} \label{sec:eva}
 To evaluate the prediction performance, 
 the RMSE for failure mode $k$ is calculated as
 \begin{align*}
 	\textrm{RMSE}_k(t) = \frac{1}{n_{\textrm{test}}}\sum_{i = 1}^{n_{\textrm{test}}}   
 		\left\{\frac{1}{N}\sum_{j = 1}^N    \left[ \hat F_j(t, k \mid \bx_i) - F(t,k \mid \bx_i) \right]^2 \right \}^{1/2} ,
 \end{align*}
 where $n_{\textrm{test}}$ is the number of units in the test set, $N=100$ is the total number of simulated datasets under each simulation scenario, $\hat F_j(t, k\mid \bx_i)$  is the predicted CIF, and  $F(t,k\mid \bx_i)$ is the true CIF. The predicted CIFs for neural networks are calculated according to formulas in Appendix~\ref{apd2}. To show the performance of SSH-Net compared to NFG and DeepHit, the ratio of the RMSEs are calculated, which are,
  $$\textrm{Ratio}_{\textrm{SSH-NFG,k}}(t) = \frac{\textrm{RMSE}_{\textrm{SSH},k}(t)}{ \textrm{RMSE}_{\textrm{NFG},k}(t)}, \; \textrm{and} \; \textrm{Ratio}_{\textrm{SSH-DeepHit,k}}(t) = \frac{\textrm{RMSE}_{\textrm{SSH},k}(t)}{ \textrm{RMSE}_{\textrm{DeepHit},k}(t)}.$$ A ratio smaller than 1 indicates the SSH-Net has better performance in terms of RMSE.
 
  Additionally, the commonly used time-dependent weighted Brier score and time-dependent AUC are calculated. The Brier score is obtained based on  \shortciteN{gerds2006consistent}, and is used to estimate the mean squared error (MSE) in predicting CIF for right-censored data when truth is unknown. 
  In particular, for one simulated dataset, the weighted Brier score for failure type $k$ at time $t$ is calculated as, 
 \begin{align*}
\mathrm{Brier}_k(t)
  = \frac{1}{n_{\textrm{test}}} \sum_{i=1}^{n_{\textrm{test}}}
  \left\{ \mathbbm{1}\{t_i> t ^{} \}  -  \left[1 - \hat F(t , k \mid \bx_i ) \right] \right \}^2
  \, \omega(t_i, t ),
  \end{align*}
  where \[
  \omega(t_i, t )
  = \frac{\mathbbm{1}\{t_i \leq t  \} \, \delta_{ik}}
  {\hat{S}_c( t_{i}^- )}
  + \frac{\mathbbm{1}\{t_i > t \}}
  {\hat{S}_c(t)},
  \]
  and $\hat{S}_c( t )$ is the Kaplan-Meier estimate for the survival function of the censoring time $C_i$ based on the test set. The weighted AUC is calculated based on \shortciteN{blanche2013estimating}, and can be used to evaluate the concordance of the ranking between predicted failure times and the observed times. Let $r_i = \hat F(t, k\mid \bx_i) $, $r_j = \hat F(t, k \mid \bx_j) $ , and $\hat w(t) = 1/{\hat F_c(t)}$, the weighted AUC for failure type $k$ at time $t$ is calculated as 
 \begin{align*}
&\textrm{AUC}_k(t) = \\
&\frac{
	\displaystyle
	\sum_{i=1}^{n_{\textrm{test}} } \sum_{j=1}^{n_{\textrm{test}} }
	\mathbbm{1}\left\{t_i \le t, \delta_{ik} = 1\right\}\,
	\hat{w}(t_i)
	\left[
	\mathbbm{1}\{t_j > t\}\,\hat{w}(t)
	+ 
	\mathbbm{1}\left \{t_j \le t, \delta_{ik}=0, \sum_{k' = 1}^K\delta_{ik'} > 0\right \}\,
	\hat{w}(t_j)
	\right]
	\mathbbm{1}\{r_i > r_j\}
}{
	\displaystyle
	\left(
	\sum_{i=1}^{n_{\textrm{test}} }
	\mathbbm{1}\{t_i \le t, \delta_{ik}= 1\}\,
	\hat{w}(t_i)
	\right)	\left\{
	\sum_{j=1}^{n_{\textrm{test}} } 
	\left[
	\mathbbm{1}\{t_j > t\}\,\hat{w}(t)
	+
	\mathbbm{1}\left\{t_j \le t, \delta_{ik}=0, \sum_{k'=1}^K \delta_{ik'}> 0 \right \}\,
	\hat{w}(t_j)
	\right]
	\right\}
}.
\end{align*}
 The Brier and AUC are evaluated on a time grid 
$(0.3, 0.4, \ldots, 0.9, 0.95) \times \textrm{max}\left(\bt_{\textrm{test}} \right)$, where $\bt_{\textrm{test}}$ consists of all the observed times in the test set.

\subsection{Results}
The SSH-Net and NFG are applied for all the eight simulation scenarios, and DeepHit is applied only to scenarios with shut down date January 1, 2020, and January 1, 2018, because of its longer  computation time.

\begin{figure}
	\centering
	\begin{tabular}{cc}
		
		\multicolumn{2}{c}{	\includegraphics[width = 0.8\textwidth]{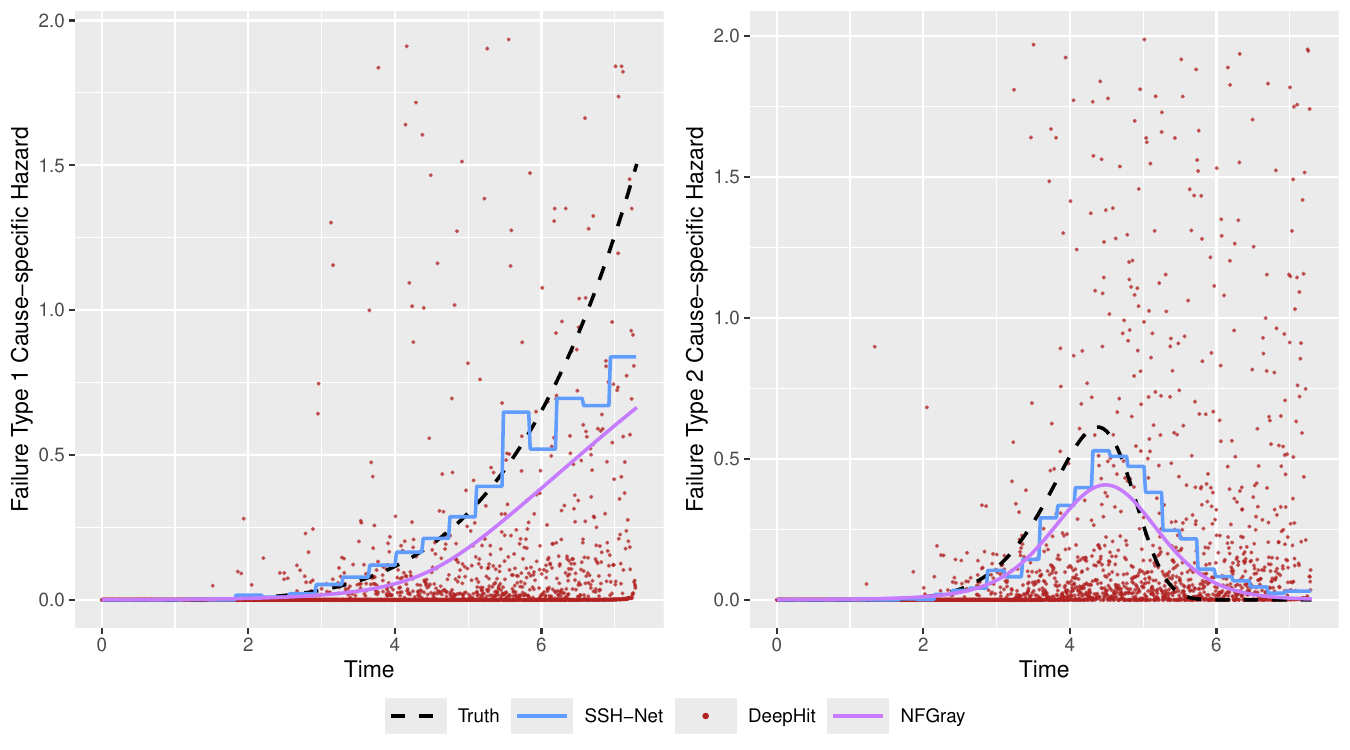}}
\\
\quad \quad \quad \quad \quad \quad(a) Failure Type 1  & \quad\quad (b) Failure Type 2 \\
		\multicolumn{2}{c}{	\includegraphics[width = 0.8\textwidth]{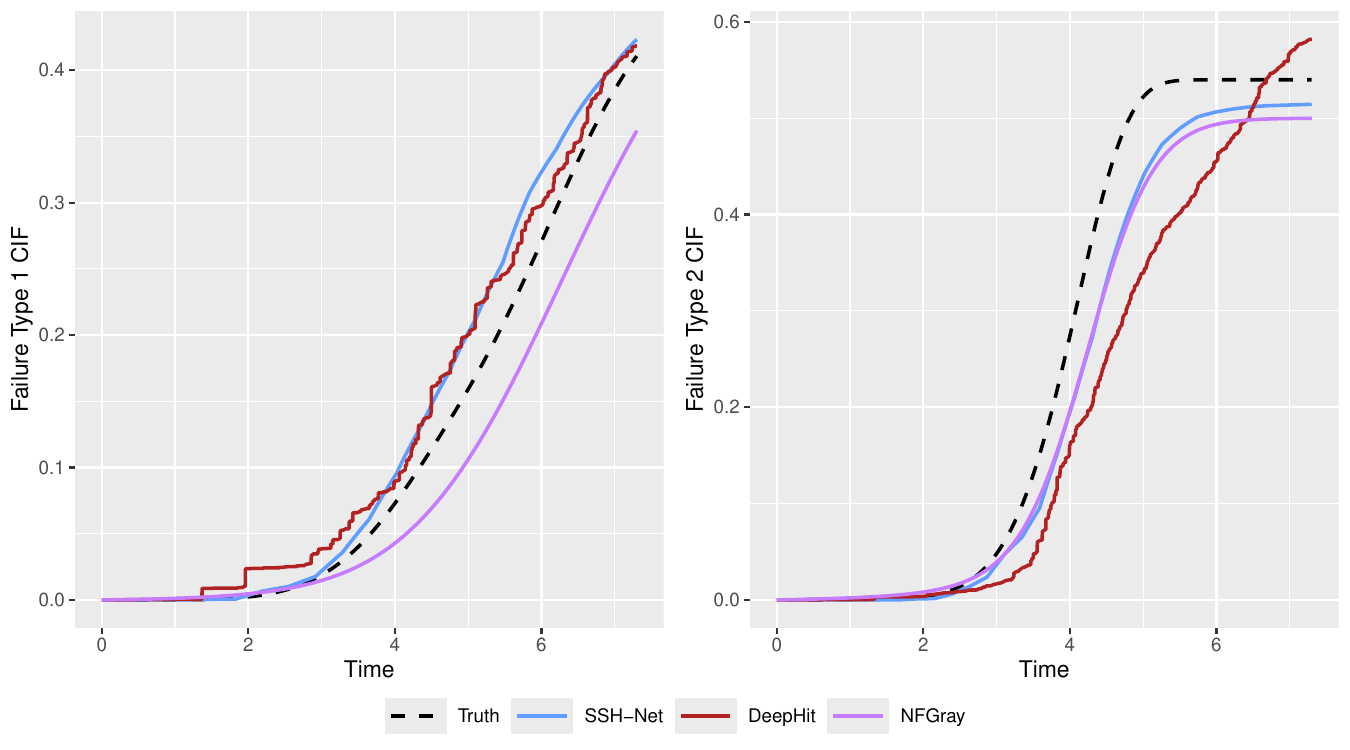}} \\
\quad \quad \quad \quad \quad \quad(c) Failure Type 1 & \quad \quad(d) Failure Type 2 \
	\end{tabular}
	\caption{Predicted cause-specific hazards and CIFs for one test unit based on one simulated dataset with $n = 10,000$ and shut down date January 1, 2020. } \label{fig:simre1}
\end{figure}

Figure~\ref{fig:simre1} shows the prediction for one test unit from a simulated dataset with $n  = 10,000$ and shut down date January 1, 2020. The cause-specific hazard predictions are illustrated in Figure~\ref{fig:simre1}(a)(b), and the CIF predictions are in  Figure~\ref{fig:simre1}(c)(d). Because of the piecewise hazard assumption, the predicted hazard function from SSH-Net is a step function. The smooth penalty in the loss function prevents wiggly behavior of the hazard functions.  Both SSH-Net and NFG successfully learn the trend of the hazard functions, while DeepHit fails to show the trend because it discretizes the failure time and predicts the failure probabilities at every discrete time point. For Failure Type 1, the prediction from NFG is better at the tail part. For Failure Type 2, SSH-Net predicts the mode of the hazard function more accurately. Additionally, DeepHit predicts the Failure Type 1 CIF more accurately compared to the Failure Type 2 CIF.

\begin{figure}
	\centering
	\begin{tabular}{cc}
		\includegraphics[width = 0.48\textwidth]{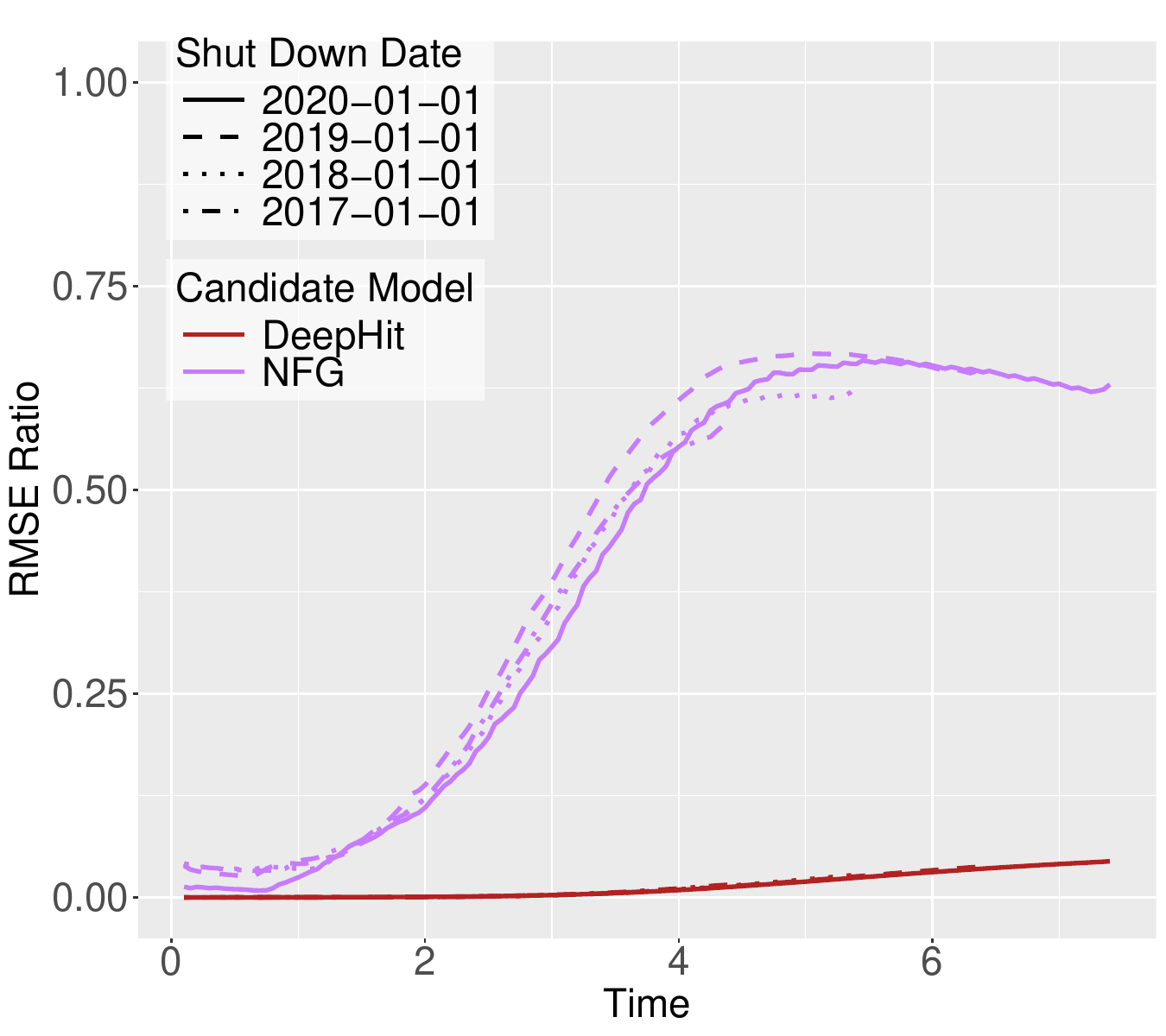} & 
		\includegraphics[width = 0.48\textwidth]{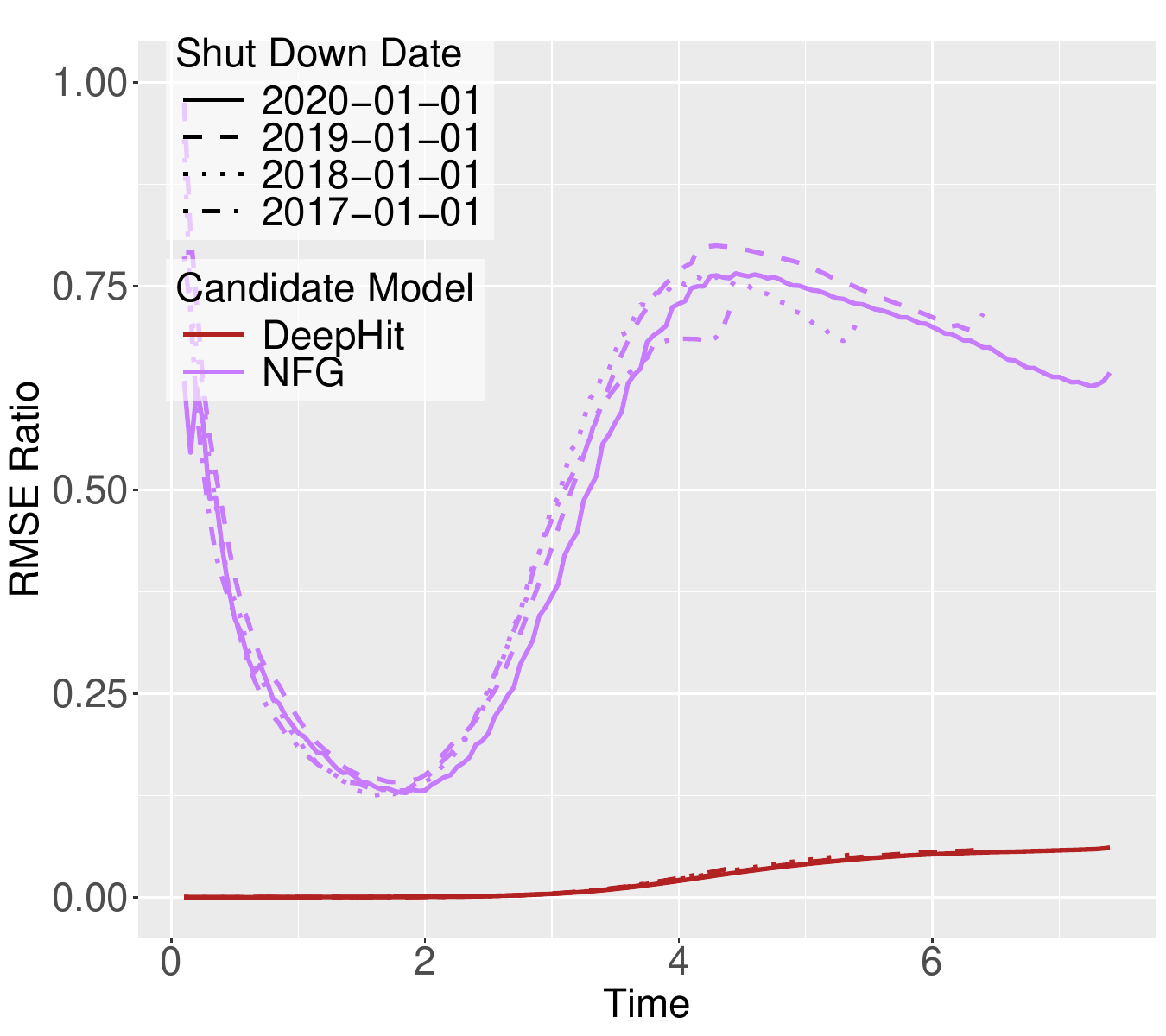}\\
		(a) Failure Type 1, $n = 10,000$ & (b) Failure Type 2, $n = 10,000$\\
		\includegraphics[width = 0.48\textwidth]{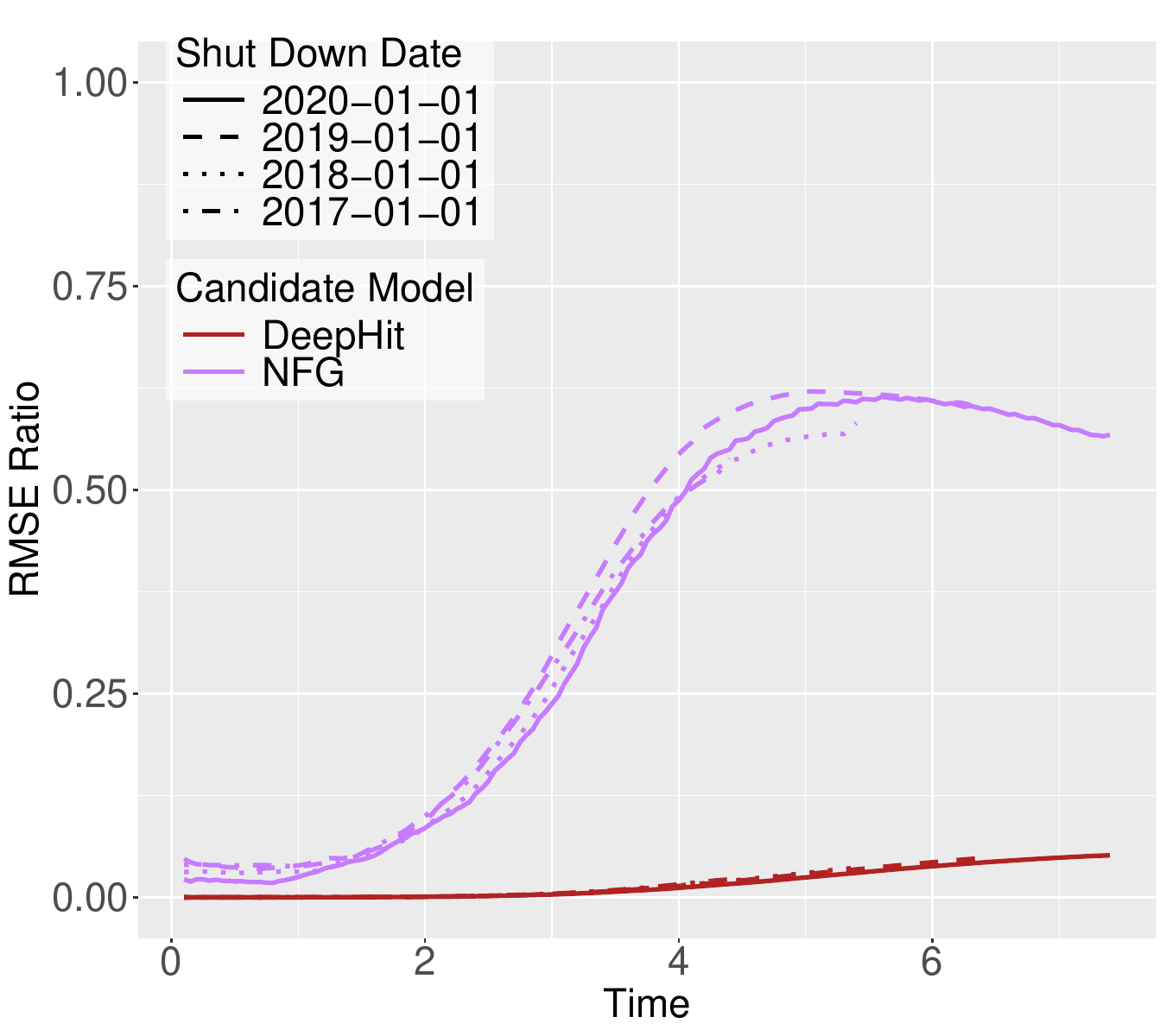} & 
		\includegraphics[width = 0.48\textwidth]{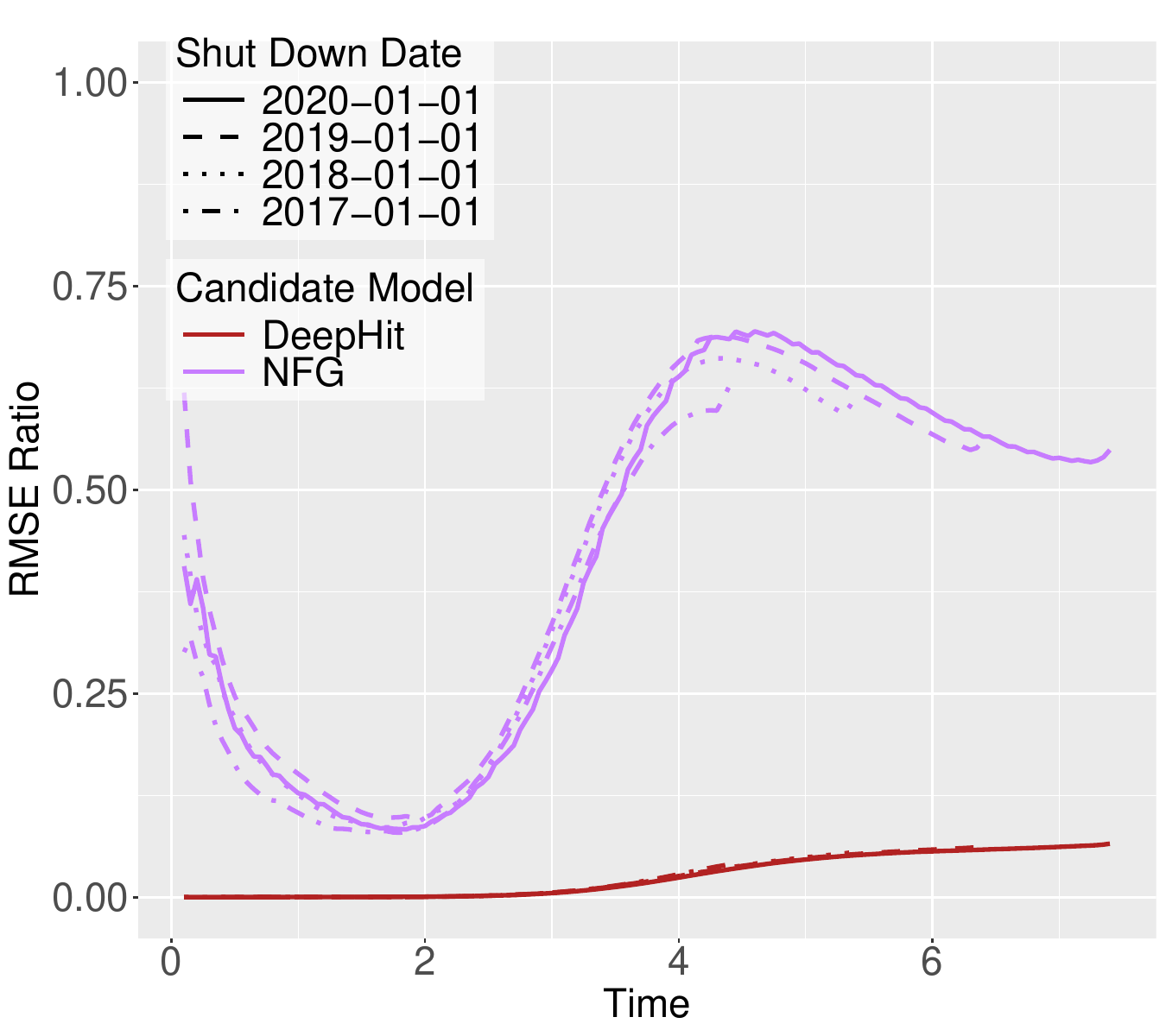} \\
		(c) Failure Type 1, $n = 5,000$ & (d) Failure Type 2, $n = 5,000$
	\end{tabular}

	\caption{The SSH-NFG RMSE ratios and SSH-DeepHit RMSE ratios across all simulation scenarios. A ratio smaller than 1 indicates the proposed SSH-Net has better performance. } \label{fig:simre3}
\end{figure}

\begin{figure}
	\centering
	\begin{tabular}{cc}
		\includegraphics[width = 0.48\textwidth]{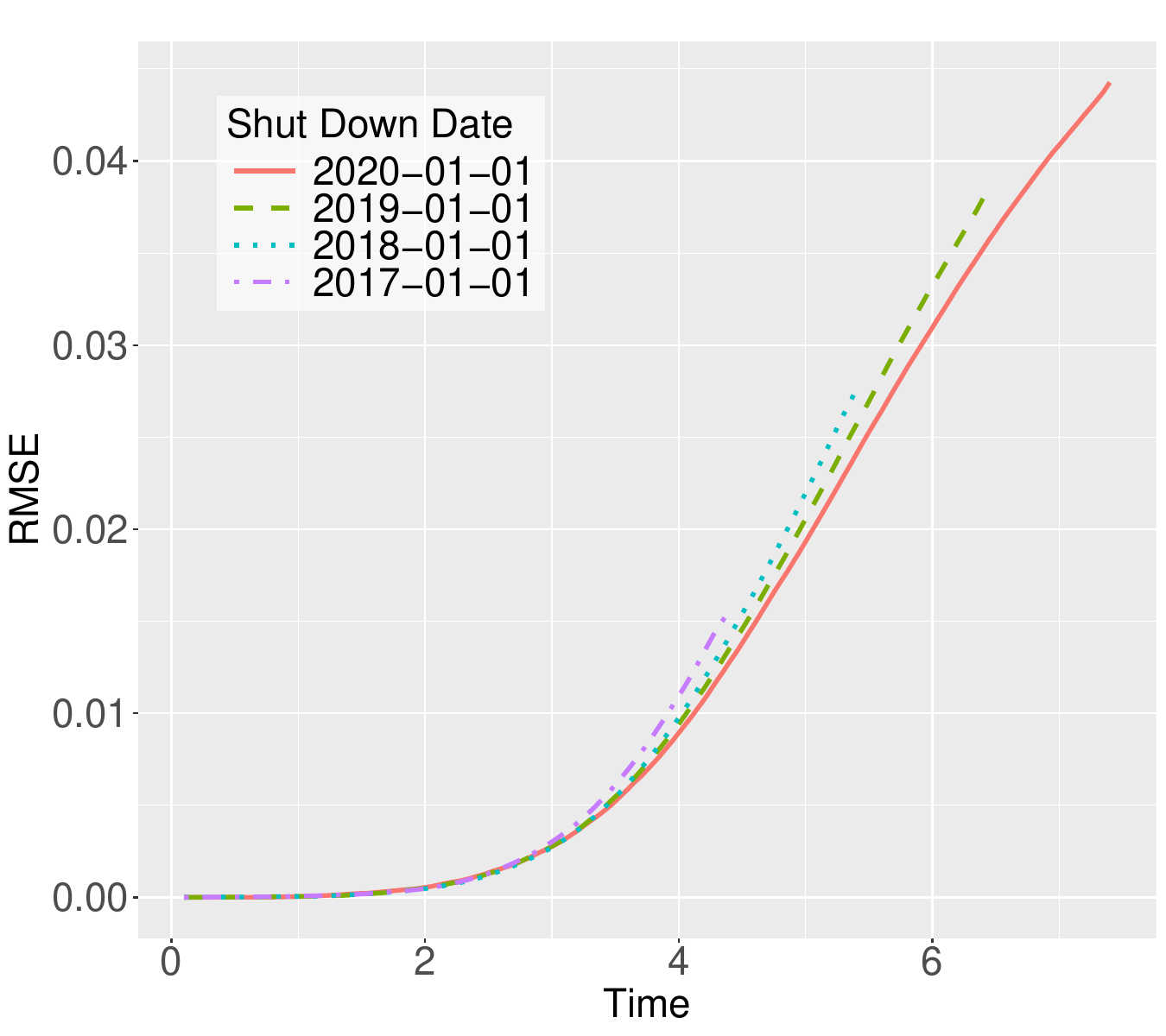} & 
		\includegraphics[width = 0.48\textwidth]{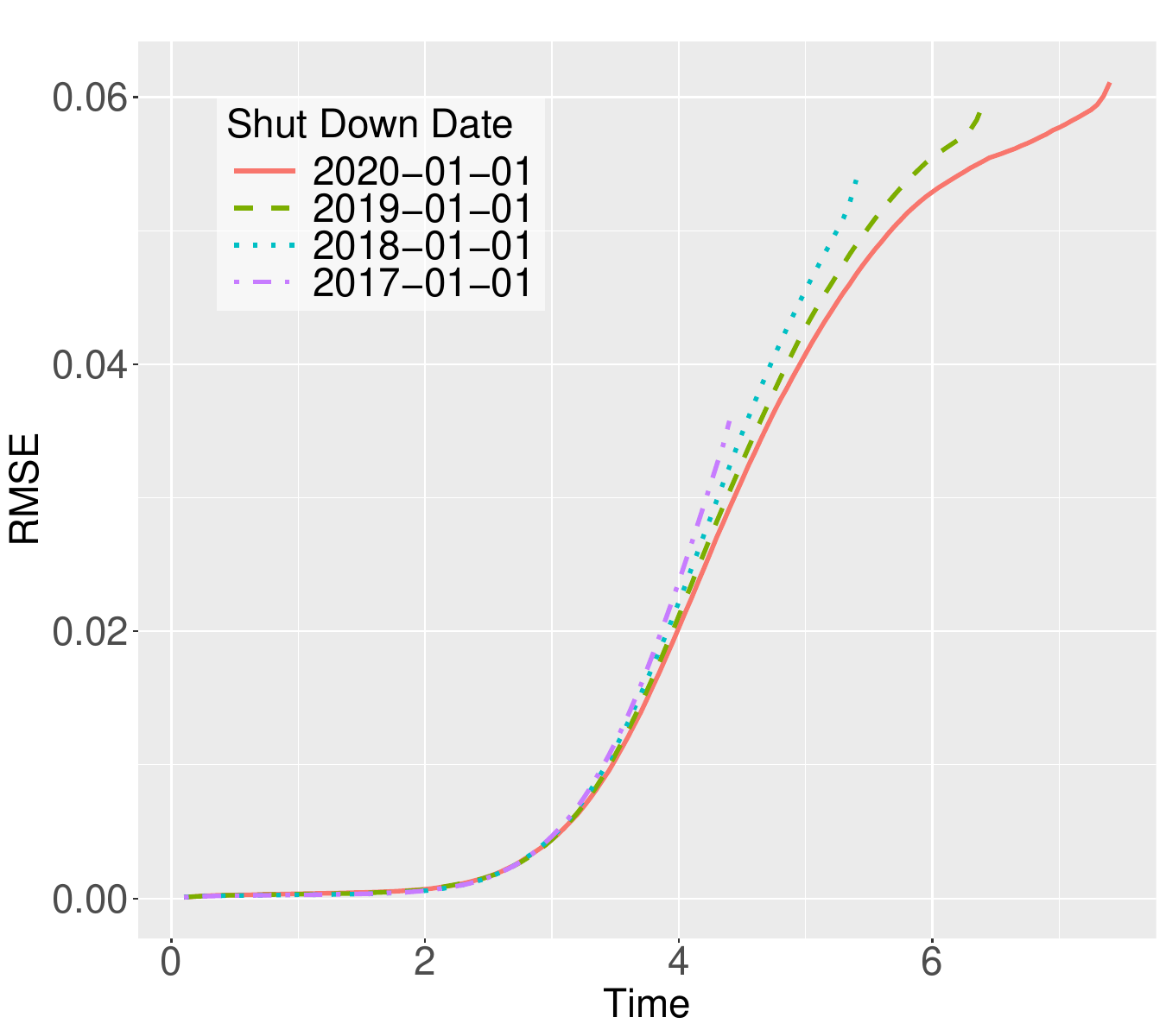}\\
		(a) Failure Type 1, $n = 10,000$ & (b) Failure Type 2, $n = 10,000$\\
		\includegraphics[width = 0.48\textwidth]{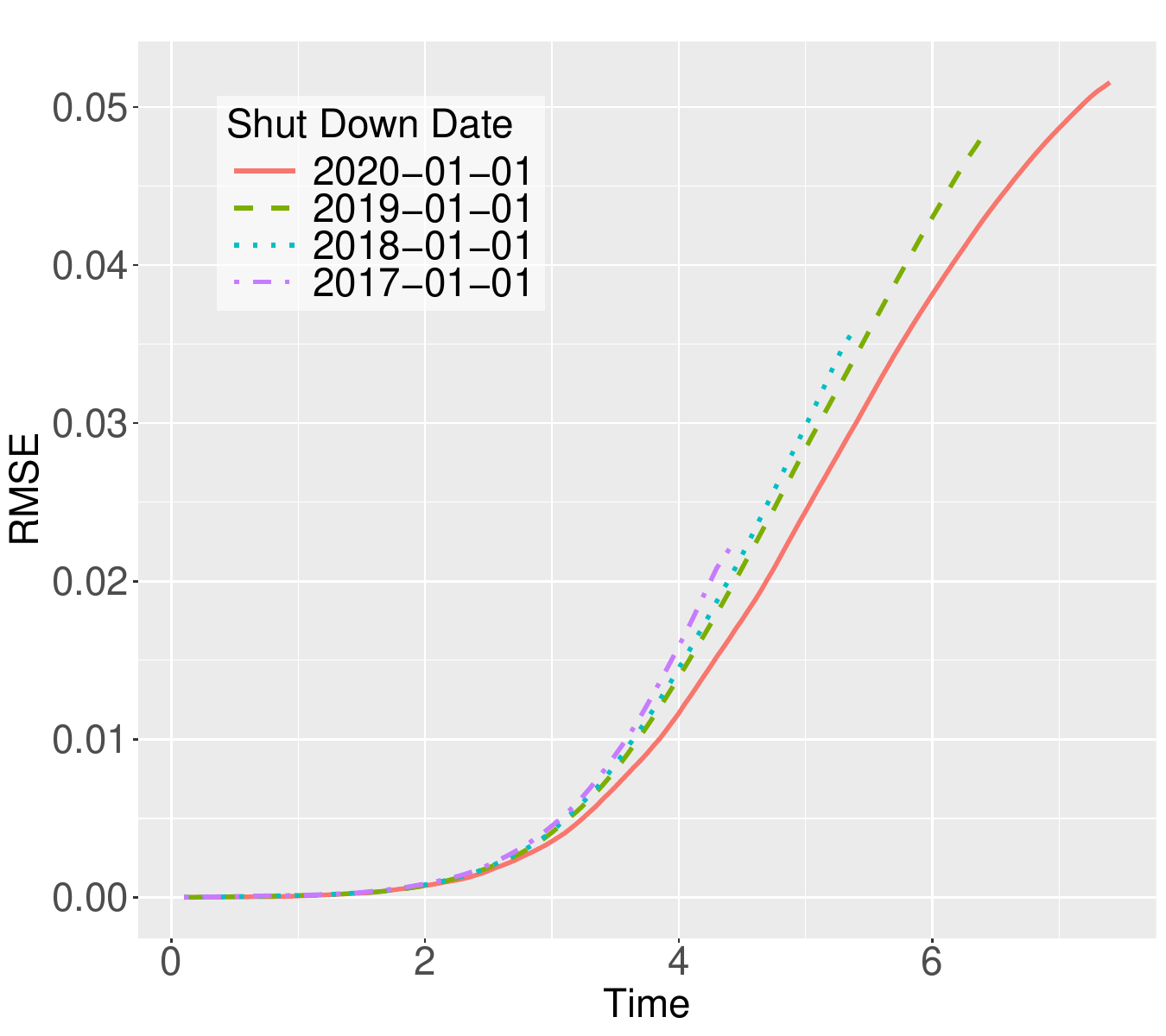} & 
		\includegraphics[width = 0.48\textwidth]{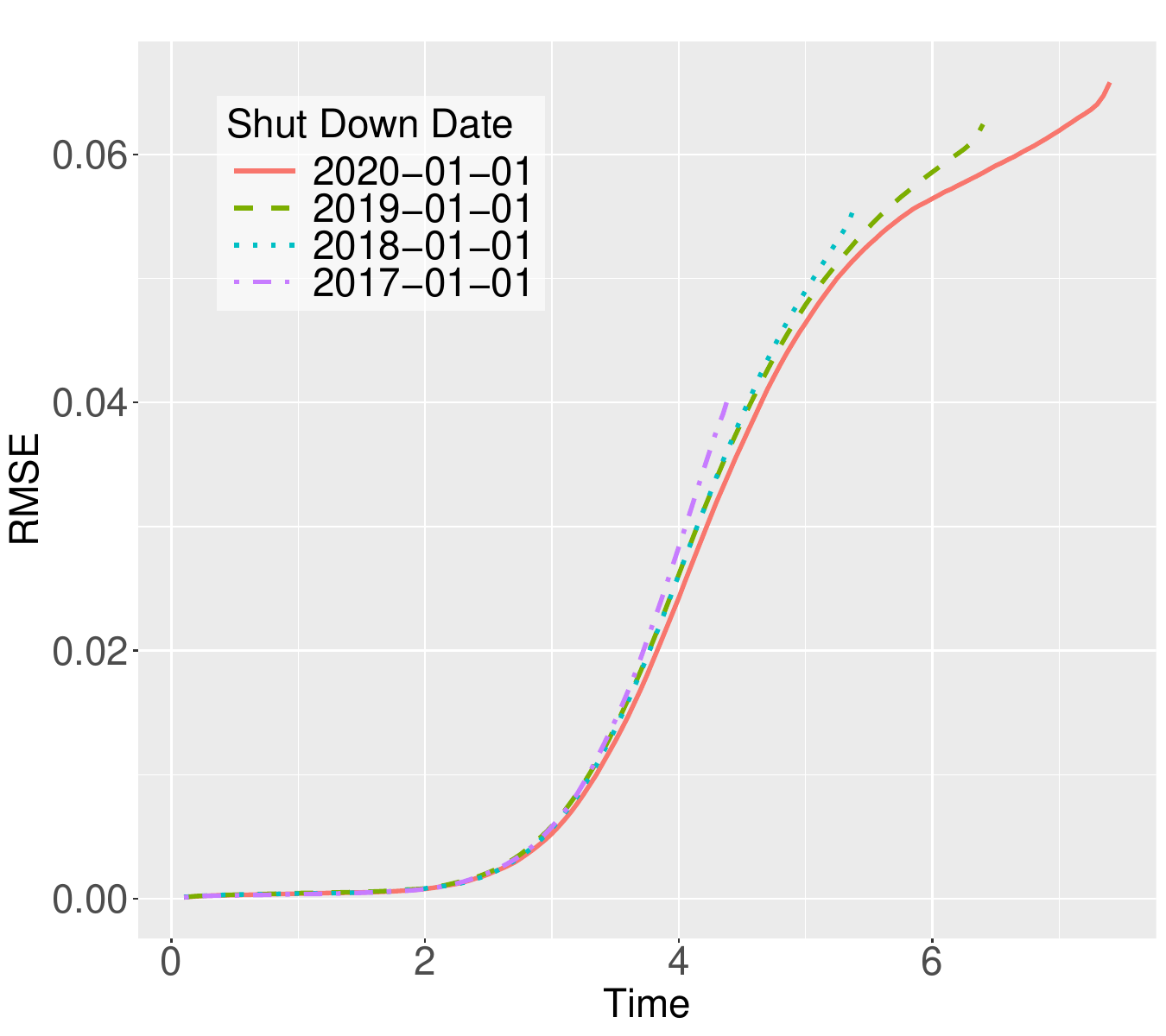} \\
		(c) Failure Type 1, $n = 5,000$ & (d) Failure Type 2, $n = 5,000$
	\end{tabular}

	\caption{RMSE of SSH-Net calculated across multiple simulation scenarios. } \label{fig:simre2}
\end{figure}

Figure~\ref{fig:simre3} shows the SSH-NFG RMSE ratio and SSH-DeepHit RMSE ratio. The results indicate that SSH-Net performs better than NFG and DeepHit in terms of RMSE, as the ratio at all evaluation time points for all scenarios are below 1. Additionally, NFG has lower RMSE compared with DeepHit. The SSH-NFG RMSE ratio first increases over time and then decreases slightly, indicating that SSH-Net performs better at smaller time values. The ratio reaches the highest value around 4 years, indicating the performance of SSH and NFG are more similar around that time. Figure~\ref{fig:simre2} shows the RMSE of SSH-Net across all the simulation scenarios. The RMSE increases through time, showing that the prediction is less accurate for large time values. Overall, the RMSE are reasonably small as it is calculated based on the test data. 

\begin{table}[]
	\centering
	\caption{Brier score and AUC averaged on the evaluation time grid and across 100 simulated datasets. The smallest Brier score and the largest AUC are highlighted in bold. The scores are reported in units of 1e-2. }\label{tab:brier}
	\begin{tabular}{cc|cc|cc}
		\hline\hline
		\multicolumn{2}{c|}{Shut Down Date}  & 2020-01-01  &  2018-01-01 & 2020-01-01 & 2018-01-01\\
		\hline
		& & \multicolumn{2}{c|}{$n$ = 10,000} &  \multicolumn{2}{c}{$n$ = 5,000}\\
		\hline
		Failure Type 1 &	SSH-Net  & \bfseries 4.65   &\bfseries  1.87 &\bfseries   5.60 & \bfseries    2.32 \\
		Avg Brier (\%)	& NFG  &4.78    &  1.91 & 5.82 & 2.40\\
		& DeepHit & 5.24&2.61  & 6.43&3.20 \\ \hline
		
		Failure Type 2 &SSH-Net  & \bfseries  9.71    & \bfseries  4.38  & \bfseries  10.36 &\bfseries   4.62\\
		Avg Brier (\%)	& NFG  &9.90    & 4.46 &  10.76  & 4.79 \\ 
				& DeepHit & 13.98&  9.68& 13.70& 9.55\\\hline
		Failure Type 1 &SSH-Net  &\bfseries  82.17 & \bfseries  81.46 & \bfseries  80.19&\bfseries  81.11  \\
		Avg AUC (\%)	& NFG  &79.80    & 77.29 &75.85 & 73.91 \\
				& DeepHit & 79.39& 78.40 & 76.92 & 77.54\\
		\hline
		Failure Type 2 &	SSH-Net  & \bfseries  78.10&\bfseries   77.07 & \bfseries  77.31  & \bfseries  79.14 \\
		Avg AUC (\%)	& NFG  &77.48   & 76.38 &75.51   & 77.20 \\
				& DeepHit & 71.14& 74.34 & 70.24 & 74.64 \\
		\hline\hline
	\end{tabular}
\end{table}

 Table~\ref{tab:brier} shows the Brier and AUC  for the three models for four simulation scenarios. The scores are averaged across the evaluation time grid mentioned in Section~\ref{sec:setting}, the test set, and all $N$ simulated datasets.  The results of SSH-Net and NFG for all eight scenarios are provided in Supplementary Section 2. A lower Brier score indicates smaller MSE, and a higher AUC indicates better classification performance. For all the scenarios, SSH-Net has the lowest Brier score and highest AUC, indicating its strong performance. Note that the difference among Brier scores are small, as the CIFs are between $0$ and $1$. 
\section{Data Application}\label{sec:real}

This section introduces the usage of SSH-Net on GPU reliability analysis inside supercomputers. The Titan GPU data is introduced. In the analysis, the CIFs, together with the overall cdf of GPUs are predicted. The relative risk among multiple GPU failure types are further discussed. The results can also show the influence factors of GPU failure times.

\subsection{The Titan GPU Data}\label{sec:data}

The Cray XK7 Titan supercomputer GPU operation data published in \shortciteN{ostrouchov2020gpu} includes time-to-event data of over {30,000} GPUs collected for nearly 7-year-long period. There are two main reasons that trigger the failure of a GPU. One is the Double Bit Error (DBE), which identifies a double bit flip to prevent the use of corrupted data. The other is the Off the Bus (OTB) that flags the loss of the host CPU connection to the GPU. One GPU node can fail multiple times for various reasons during the Titan supercomputer service time. As the later failures can be triggered by the first failure (\shortciteNP{ostrouchov2020gpu}), our analysis focuses on the time to the first failure for each GPU. Specifically, the cleaned version of the dataset as in the study by \shortciteN{Min03072025} is used. In the cleaned Titan GPU data, there are  19,319 GPU  units, with 1127 OTB  failures and 3093 DBE failures, leading to $94.2\%$ OTB censor rate and $84.0 \%$ DBE censor rate. 

\begin{figure}
	\centering
	\includegraphics[width = 0.7\textwidth]{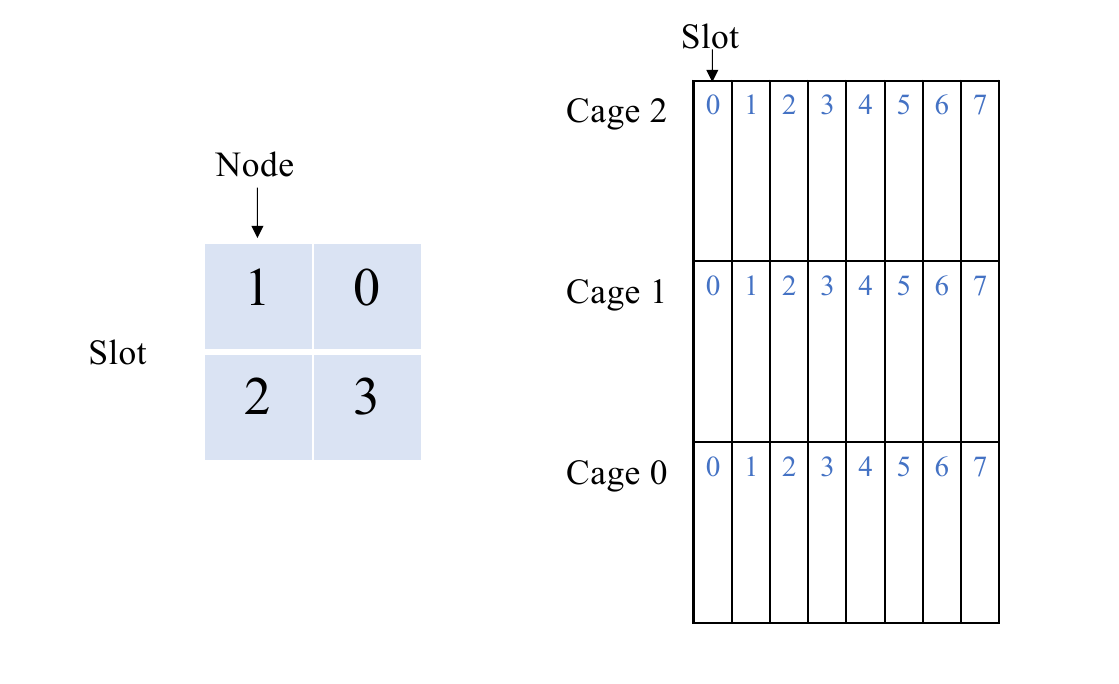}
	\caption{The hierarchical structure of Titan supercomputer within each cabinet. There are 4 GPU nodes in each slot, 8 slots in each cage and 3 cages in each cabinet.} \label{fig:titan}
\end{figure}

There are two sets of covariates that can influence the GPU failure times, and the sets are associated with different levels of the Titan supercomputer's physical structure.
Specifically, the first set of covariates represents the ``global'' spatial location of GPUs in the server room. The Titan system has 8 rows by 25 columns of cabinets, with all GPUs regularly arranged inside these 200 cabinets. Since GPU reliability has been linked to heat dissipation following the work by \shortciteN{ostrouchov2020gpu}, we use the row and column position of cabinets as the first set of covariates to indicate the spatial location of each GPU. The locations of cabinets serve as a proxy for the regional temperature affecting GPU failure.
The second set of covariates is linked to the hierarchical structure of GPUs within each cabinet. Every cabinet has a vertical stack of 3 cages, each with 8 slots. Every slot contains 4 nodes, with a single GPU installed per node. Figure~\ref{fig:titan} presents the hardware structure of GPUs within each cabinet. We consider the cage, slot, and node position of each GPU as the second set of covariates that impact GPU lifetime.

\begin{figure}
	\centering
	\includegraphics[width = 0.8\textwidth]{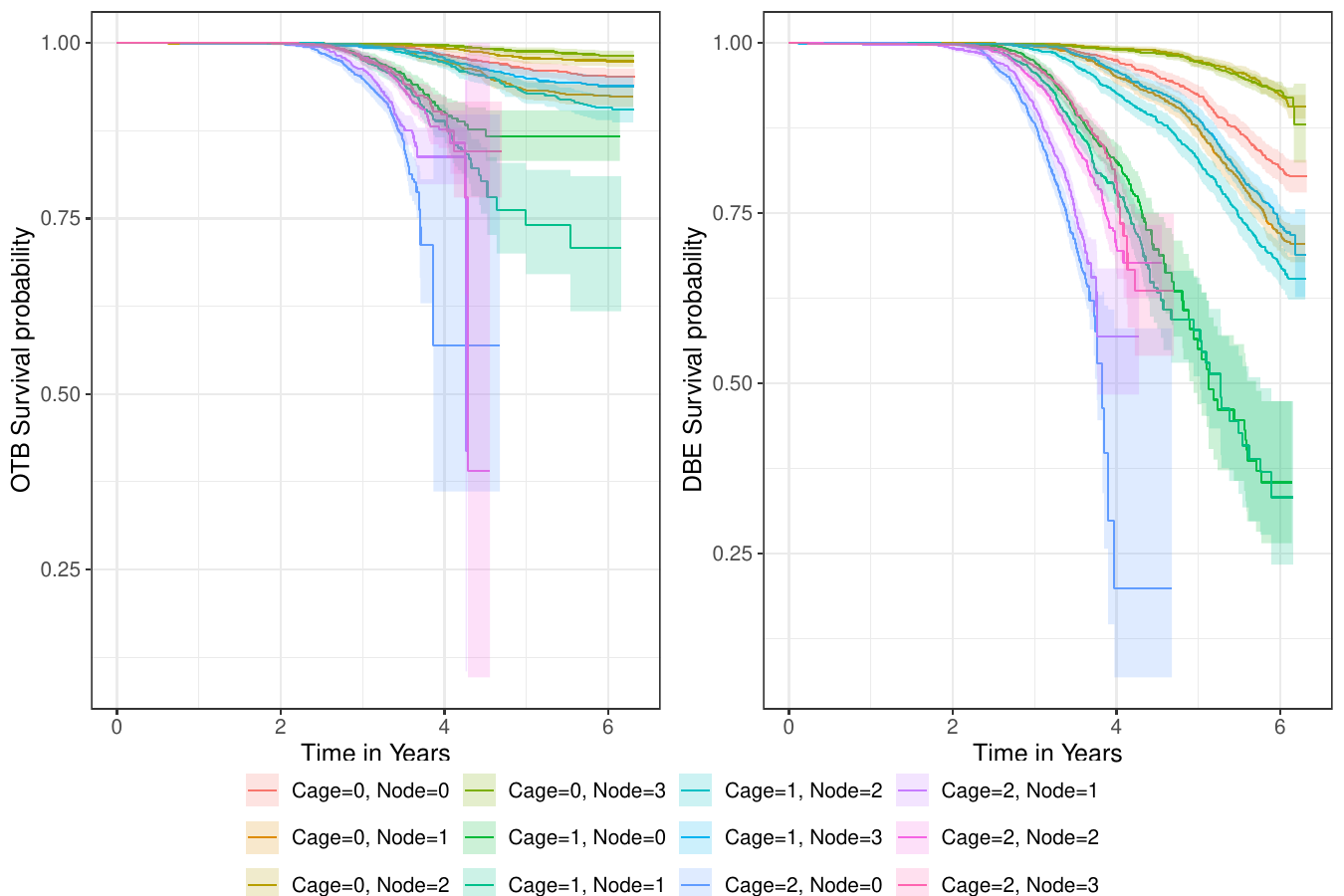}
	\caption{Kaplan-Meier estimates and the $95\%$ CIs for the OTB and DBE survival probabilities based on the cage and node covariates in the Titan GPU data.} \label{fig:gpukm}
\end{figure}

\begin{figure}
	\centering
	\includegraphics[width = 0.8\textwidth]{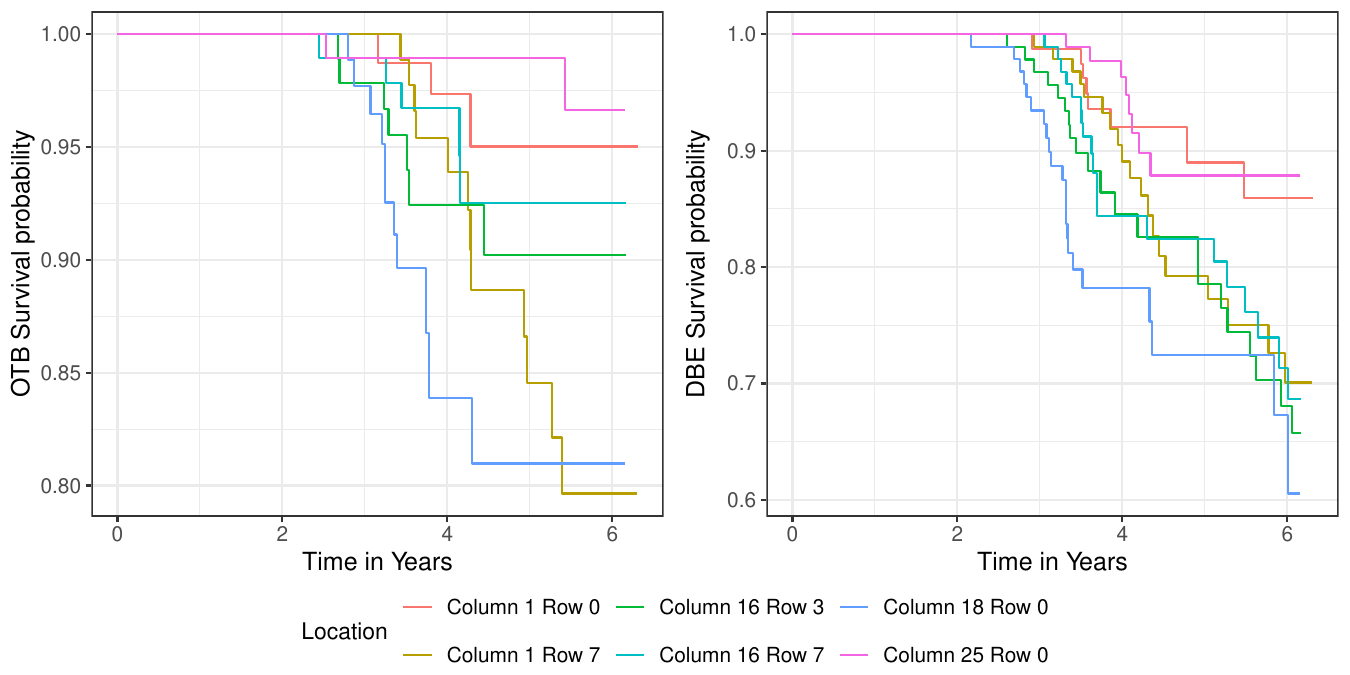}
	\caption{Kaplan-Meier estimates for the OTB and DBE survival probabilities based on the 6 different cabinet locations in the Titan GPU data.} \label{fig:gpukm2}
\end{figure}

Figure~\ref{fig:gpukm} and Figure~\ref{fig:gpukm2} visualize the Kaplan-Meier estimates of the Titan GPU data, based on cage and node positions and 6 different cabinet locations. We can see that for both OTB and DBE failures, GPUs in the top cage have a lower survival probability than those in the bottom cage. The survival probability also varies across different spatial locations. These visualizations justify the inclusion of both within cabinet hierarchies and cabinet locations as inputs to our neural network model.

\subsection{Failure Probability Prediction for the Titan GPU Data}\label{sec:dataap}

In this section, we apply the proposed SSH-Net to analyze the Titan GPU Dataset. In applying SSH-Net, we set $p_l=200$ based on the number of cabinet locations, and $p_c = 12$ based on one-hot encoding of the categorical covariates cage, slot, and node, after eliminating collinearity. To select the hyperparameters, recall that $J_k$ is related to the constant hazard assumption, and the grid can be selected based on the background of the study. For the Titan GPU data, we use a grid $\{ 10, 20, 30, 50, 70 \}$. As the Titan supercomputer operated for around $8$ years, using $70$ bins leads to the assumption that the instantaneous failure risk keeps unchanged during around 41 days, which is reasonable for a stable system. Less number of bins correspond to a longer time interval of the constant hazard. Additionally, with the smooth penalty term, the difference between adjacent hazard rates can be penalized to values close to $0$, leading to a longer interval with approximately constant hazard. To avoid overfitting, we let $J_k \leq J_{k^*}$ if $\sum_{i = 1}^n\delta_{ik} \leq \sum_{i  = 1}^n\delta_{ik^*}$. That is, the failure type with less observed failures has smaller number of bins. As suggested in Section~\ref{sec:structure}, number of nodes in $M_k$ is related to the grid of $J_k$, and $50$ nodes is used for the Titan GPU data.

\begin{figure}
	\centering
	\begin{tabular}{cc}
		\includegraphics[width = 0.48\textwidth]{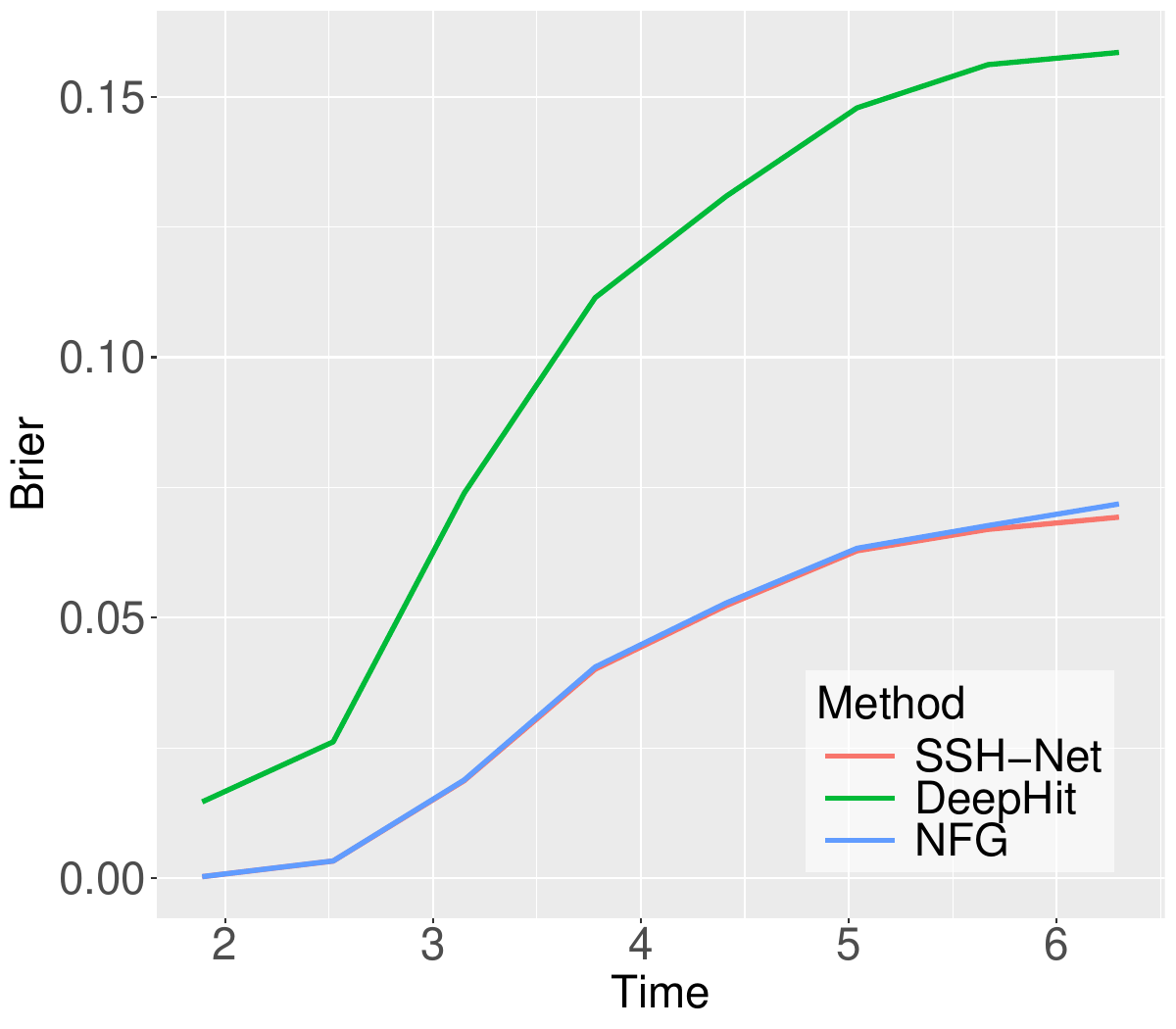} & 
		\includegraphics[width = 0.48\textwidth]{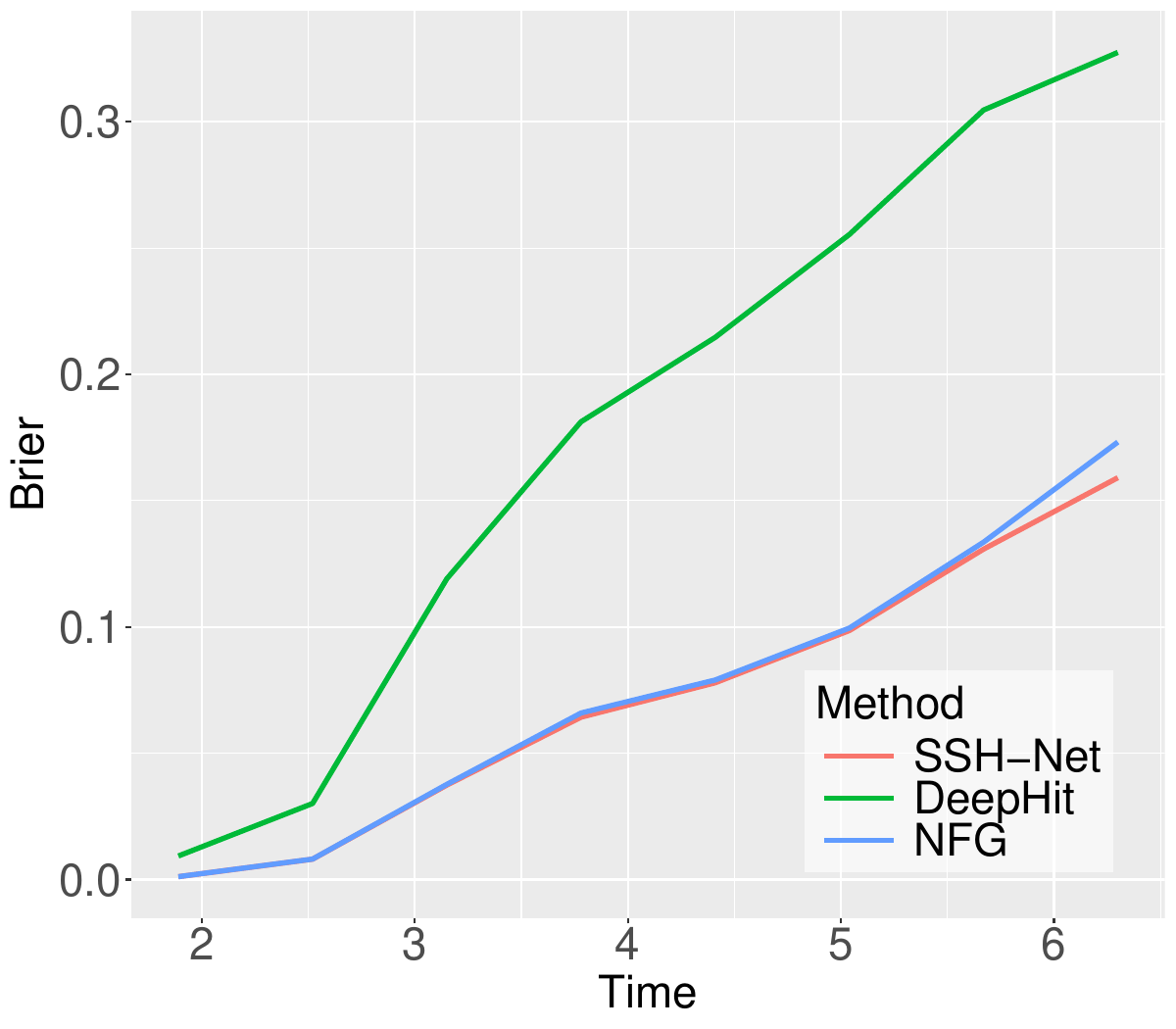}\\
		(a) OTB & (b) DBE\\
		\includegraphics[width = 0.48\textwidth]{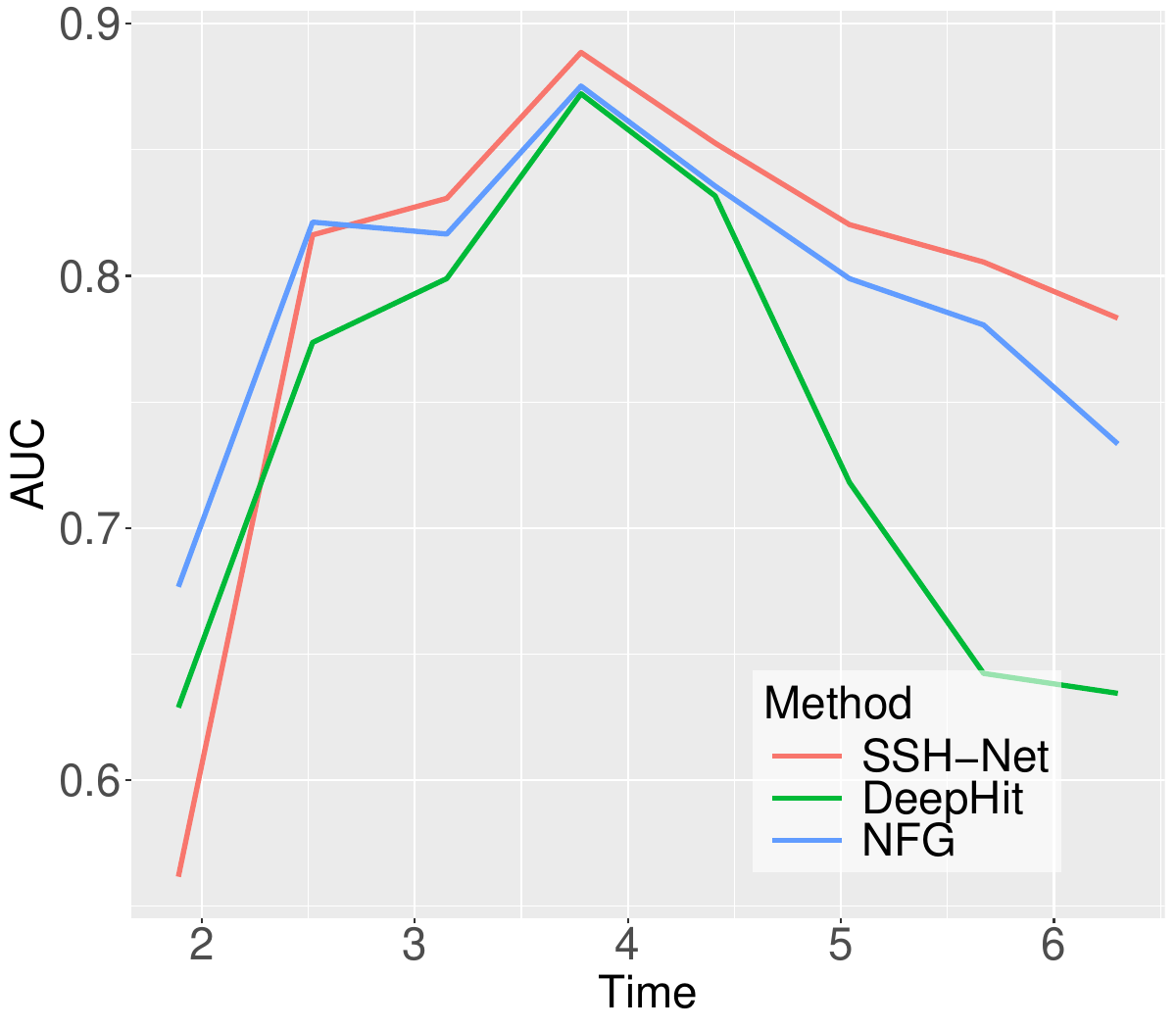} & 
		\includegraphics[width = 0.48\textwidth]{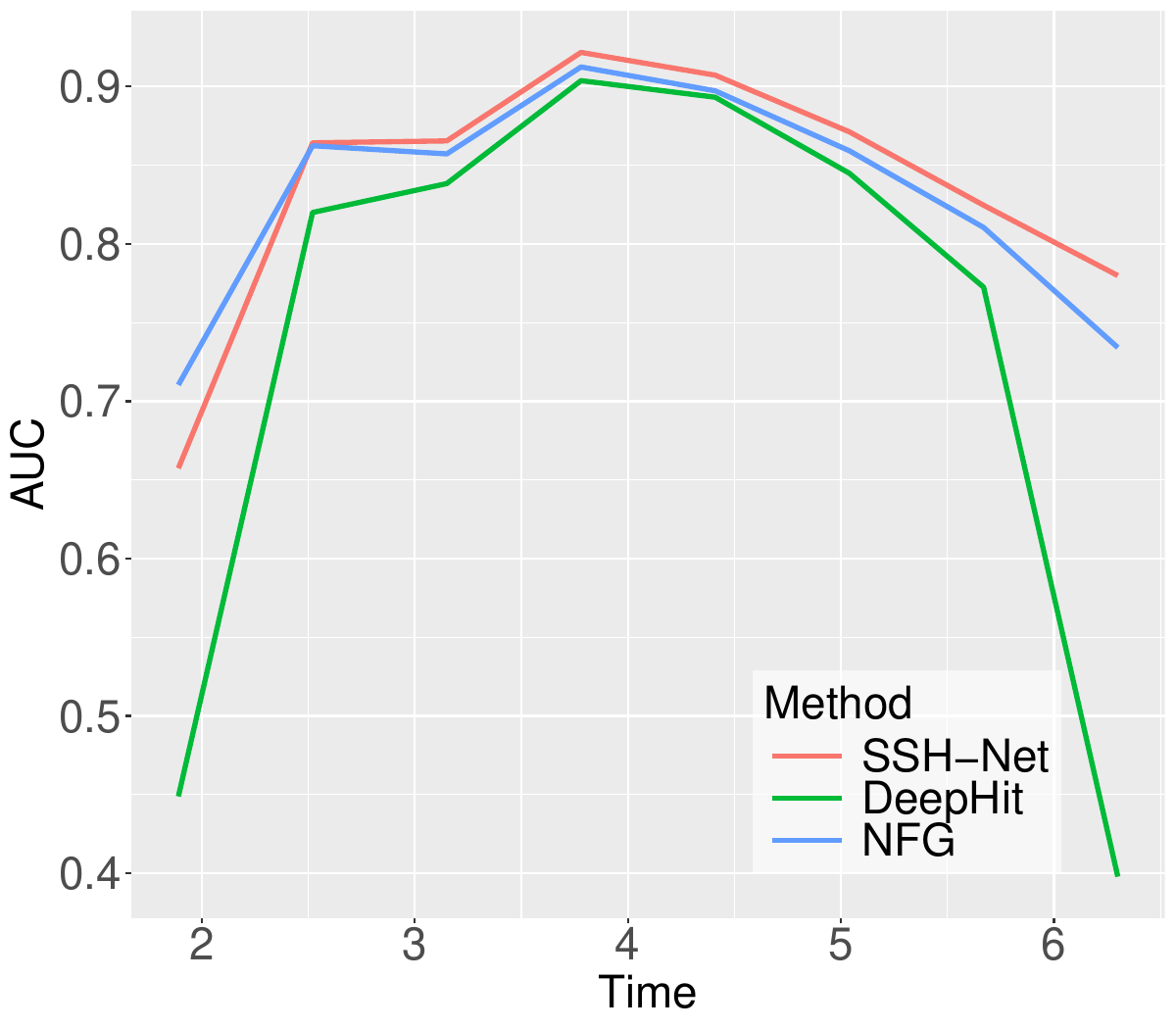} \\
		(a) OTB & (b) DBE
	\end{tabular}

	\caption{The Brier score and AUC for SSH-Net, NFG, and DeepHit, calculated based on 5-fold cross validation. } \label{fig:real2}
\end{figure}

 The NFG and DeepHit are also applied to the dataset, with hyperparameter tuning using the same grids as those in Section~\ref{sec:sim}, and the Brier score and AUC of the three neural networks are compared. 
Additionally, 5-fold cross validation is performed for model comparison. Let $ |\cdot| $ denote the number of nodes in each layer of the sub-networks, Table~\ref{tab:nnset} shows the number of nodes and layers, and other hyperparameter values in the SSH-Net selected for the 5-fold data. Dropout is applied except for the final output layer, and the dropout rate used is 0.4. Figure~\ref{fig:real2} shows the Brier score and AUC from the three models, evaluated at times presented in Section~\ref{sec:setting}. The Brier scores increase through time for all the three models, and the smallest Briers are obtained using SSH-Net. Also, SSH-Net has the highest AUC across most of the evaluation points. The result shows that SSH-Net has the best performance, which aligns with the conclusion in the simulation study. 

\begin{table}
	\caption{The selected SSH-Net structures for the 5-fold Titan GPU datasets.}\label{tab:nnset}
	\centering
	\begin{tabular}{c|ccc|ccccc}
		\hline\hline
 &	$|E|$ & $|L|$& $|S|$ &  \# of layers in $M_k$ & $J_1$ & $J_2$ & $\lambda^*_1$ & $\lambda^*_2$ \\ \hline
 Fold 1 & \multirow{5}{*}{12} & \multirow{5}{*}{200}& \multirow{5}{*}{30}& 3 & 20 & 30 & 1 & 1.5 \\
 Fold 2 & &  &  & 1 & 20 & 20 & 0 & 0 \\
 Fold 3 & & & & 1 & 20 & 20 & 0 & 0.5 \\
 Fold 4 & & & & 1 & 20 & 20 & 1.5 & 0.5 \\
 Fold 5 & & & & 1 & 20 & 30 & 0 & 1.5\\ \hline\hline
	\end{tabular}
\end{table}

\begin{figure}
	\centering
	\begin{tabular}{cc}
		\includegraphics[width = 0.48\textwidth]{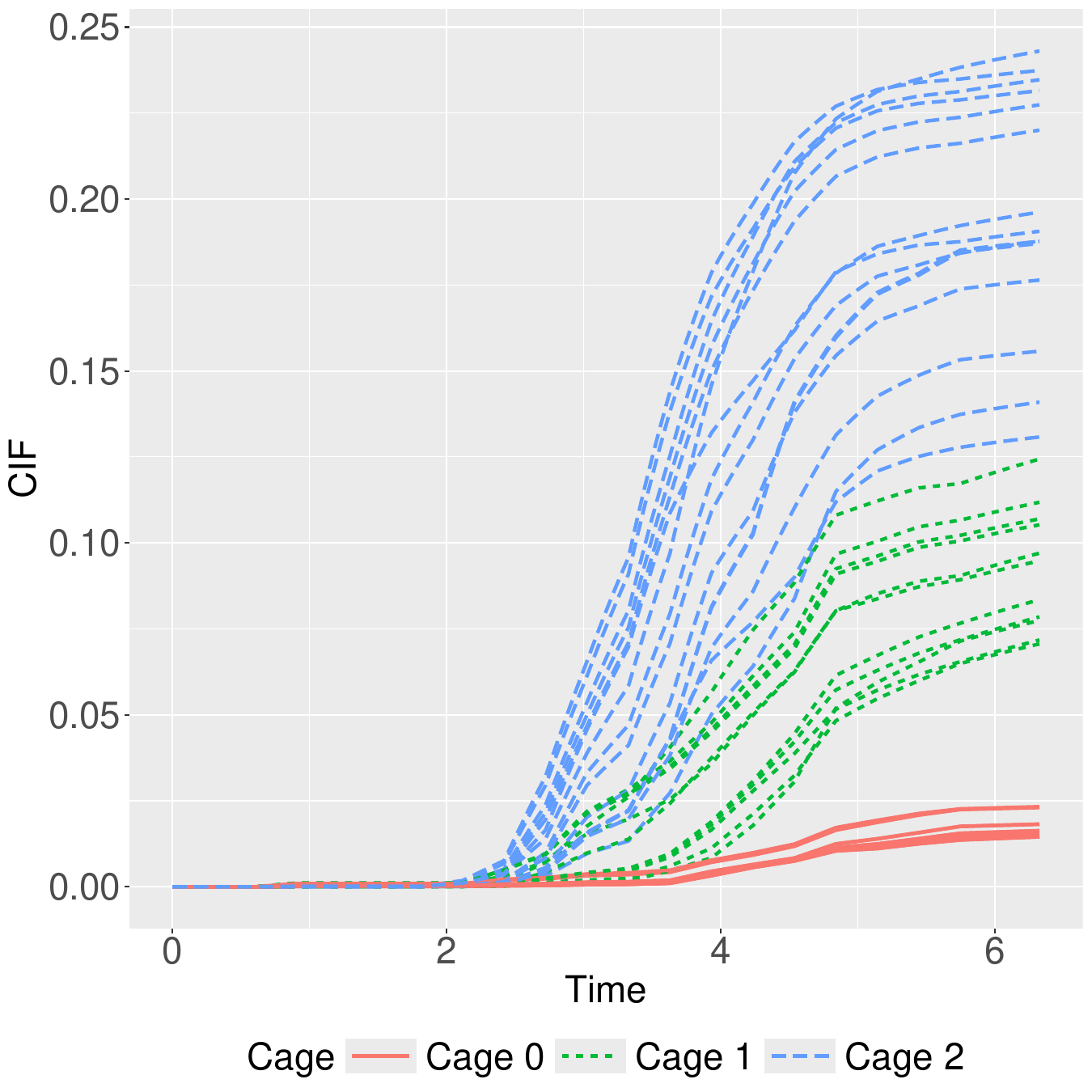} & 
		\includegraphics[width = 0.48\textwidth]{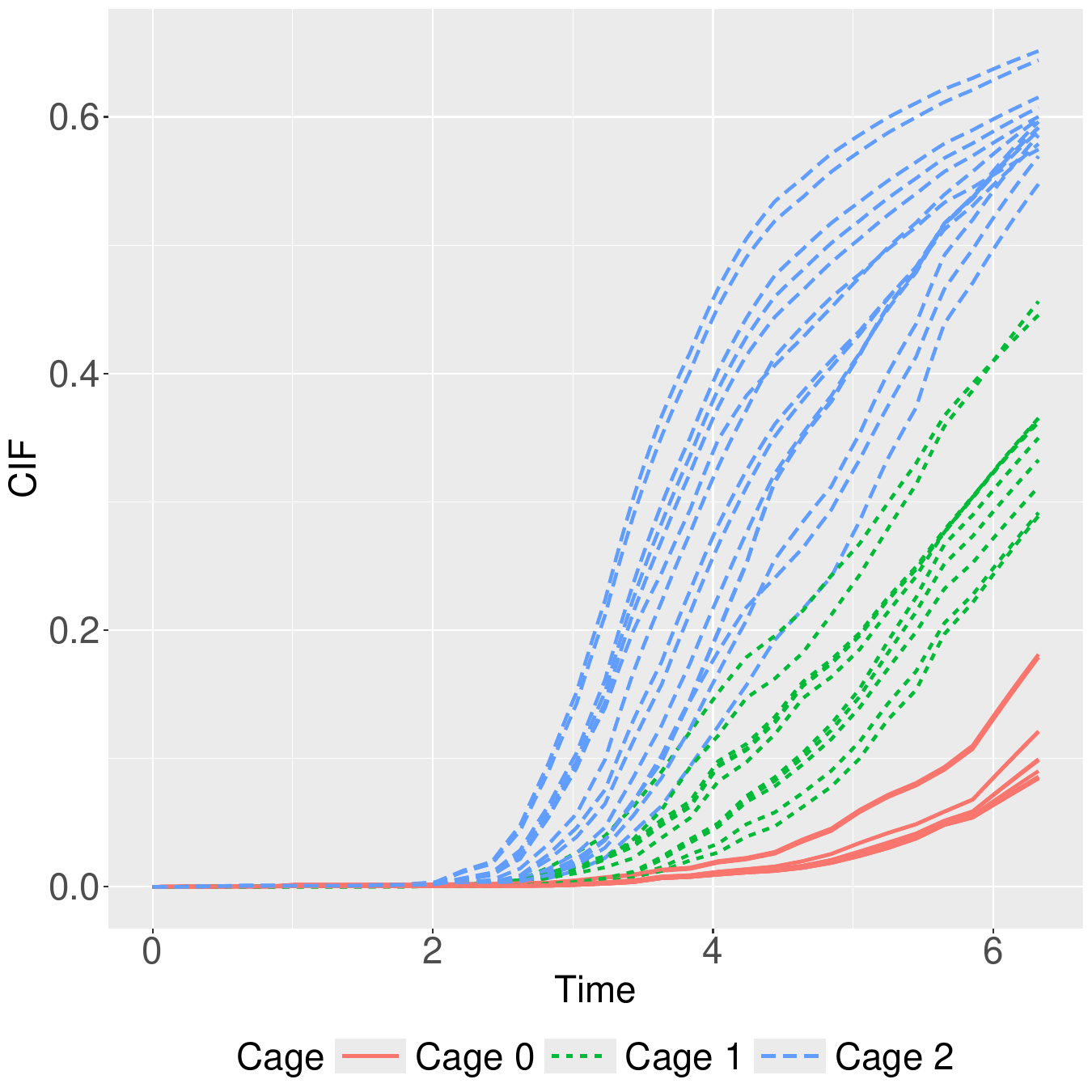}\\
		(a) OTB & (b) DBE\\
			\includegraphics[width = 0.48\textwidth]{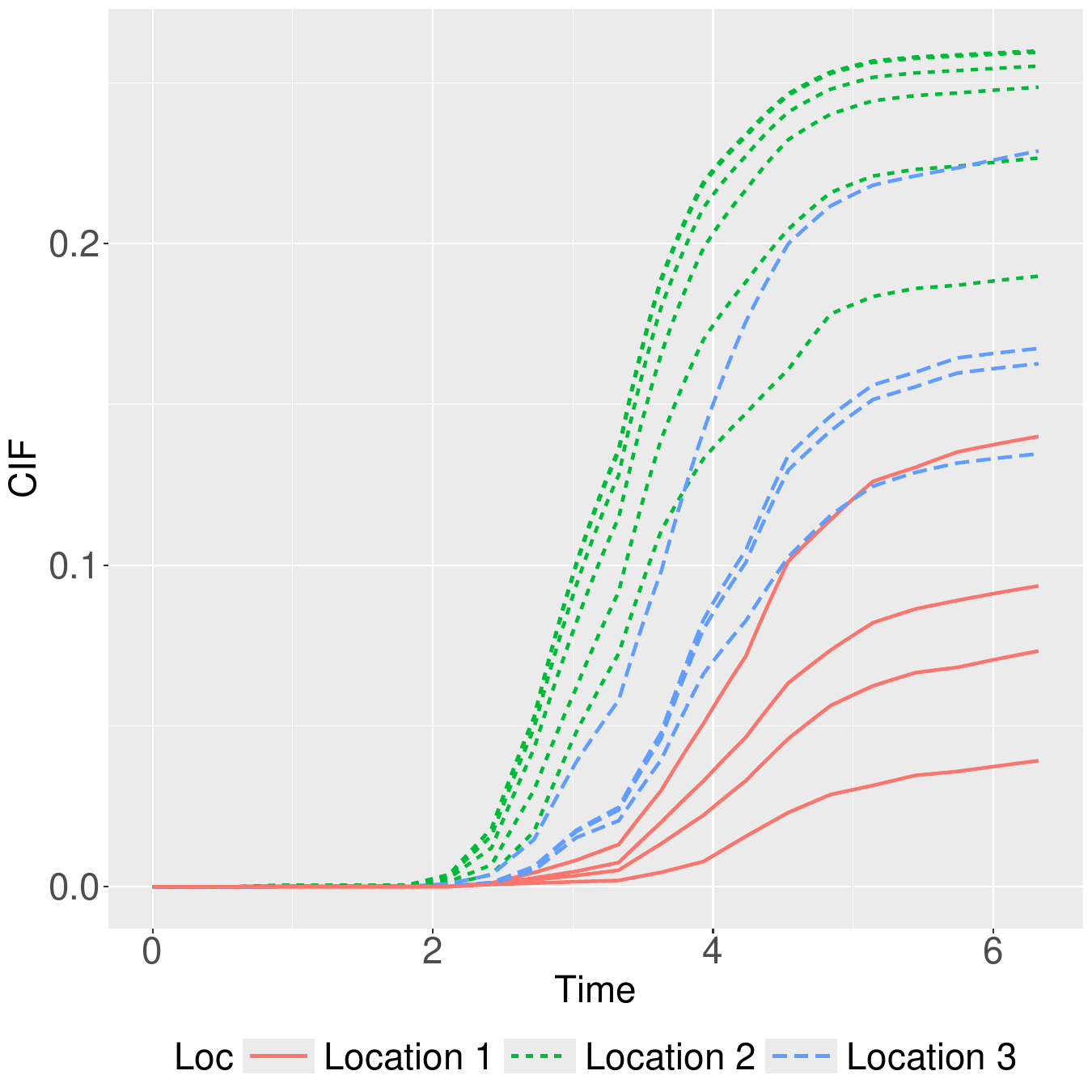} & 
		\includegraphics[width = 0.48\textwidth]{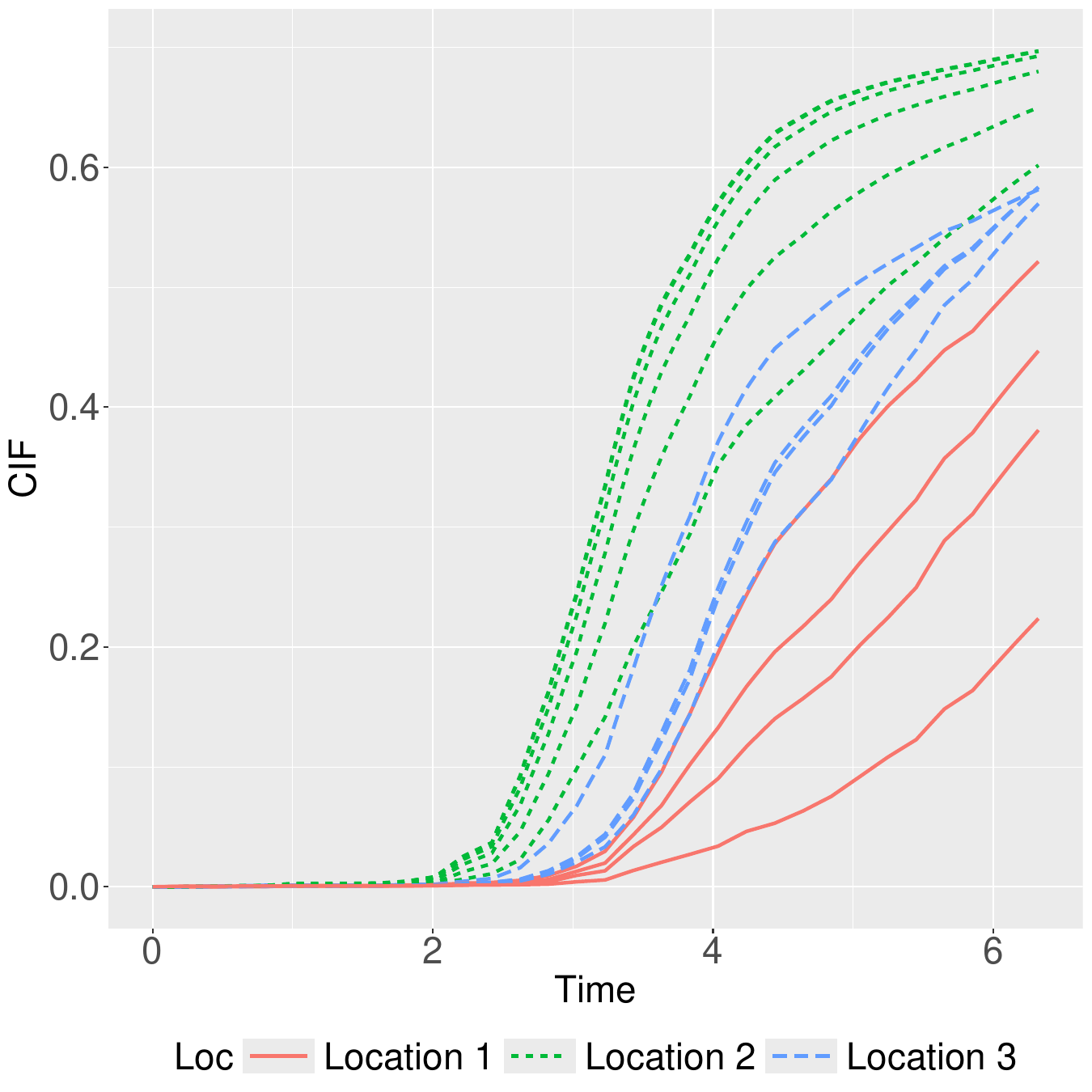}\\
		(c) OTB & (d) DBE\\
	\end{tabular}
	\caption{The predicted CIFs for OTB and DBE failures based on different cage positions and cabinet locations.}\label{fig:real3}
	\end{figure}

Figure~\ref{fig:real3} shows the predicted CIFs on different cage positions and cabinet locations for Fold 1 test data. In particular, Figure~\ref{fig:real3} (a)(b) present the predicted CIFs for GPU units at cage 0, cage 1, and cage 2,  all with node position 3 and cabinet column location 7. The slot and cabinet row locations are not fixed. The results indicate the strong influence of cage on the OTB and DBE failure probabilities. With other covariates fixed, GPUs in cage 2 have the highest failure risk, and GPUs in cage 0 have the lowest failure risk. The cage influence on GPU failure times is driven by the cooling system in the Titan's supercomputer, as introduced in \shortciteN{ostrouchov2020gpu} and \shortciteN{Min03072025}. The results indicate SSH-Net successfully captures the cage effect. Figure~\ref{fig:real3} (c)(d) present the predicted CIFs for GPUs at cabinet column 24 row 8 (location 1), column 15 row 1(location 2), and column 15 row 7 (location 3), with cage positions all fixed at 2. The slot and node positions are not fixed. The results indicate the effect of cabinet locations on the OTB and DBE failure risks. 

\begin{figure}
	\centering
	\begin{tabular}{cc}
		\includegraphics[width = 0.48\textwidth]{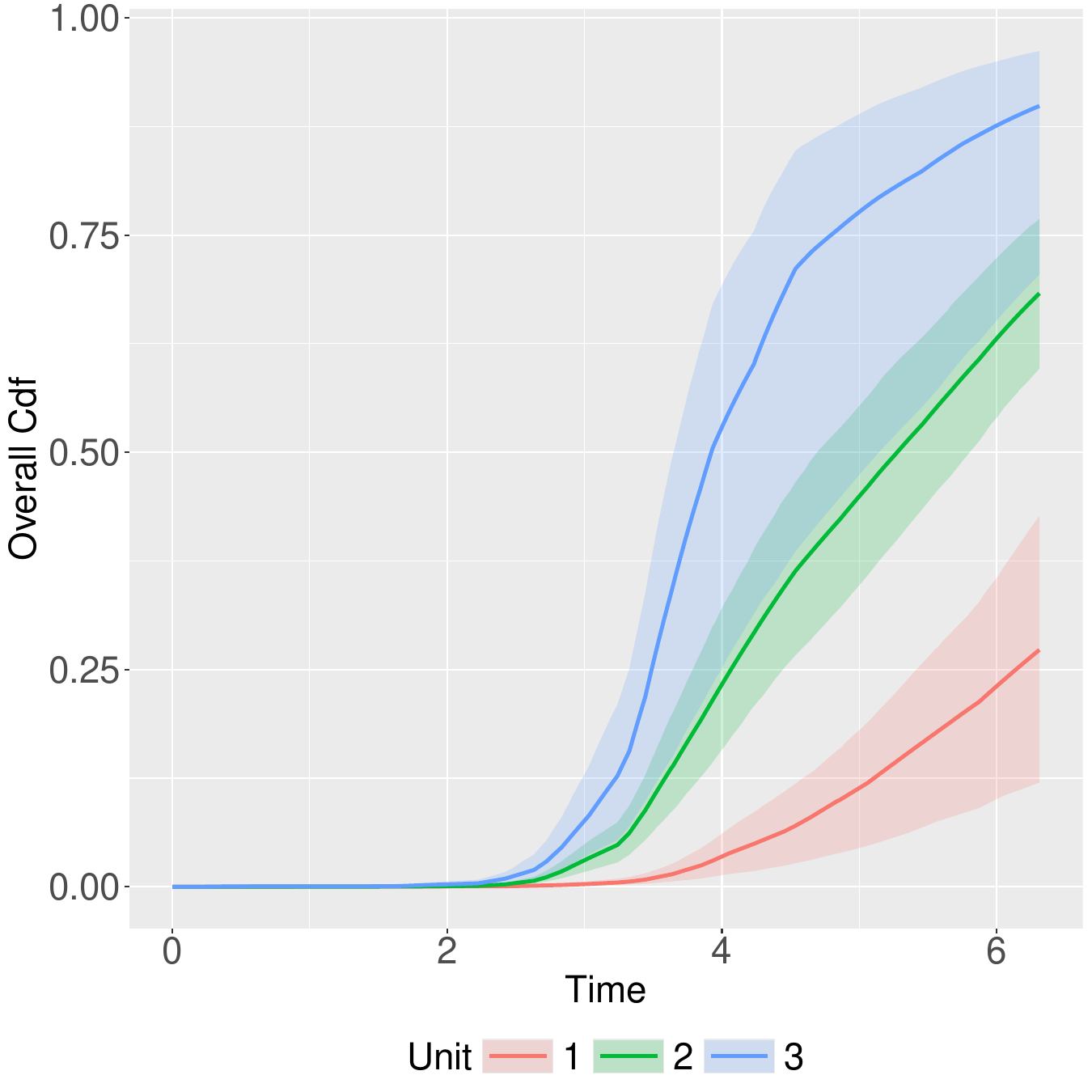} &	\includegraphics[width = 0.48\textwidth]{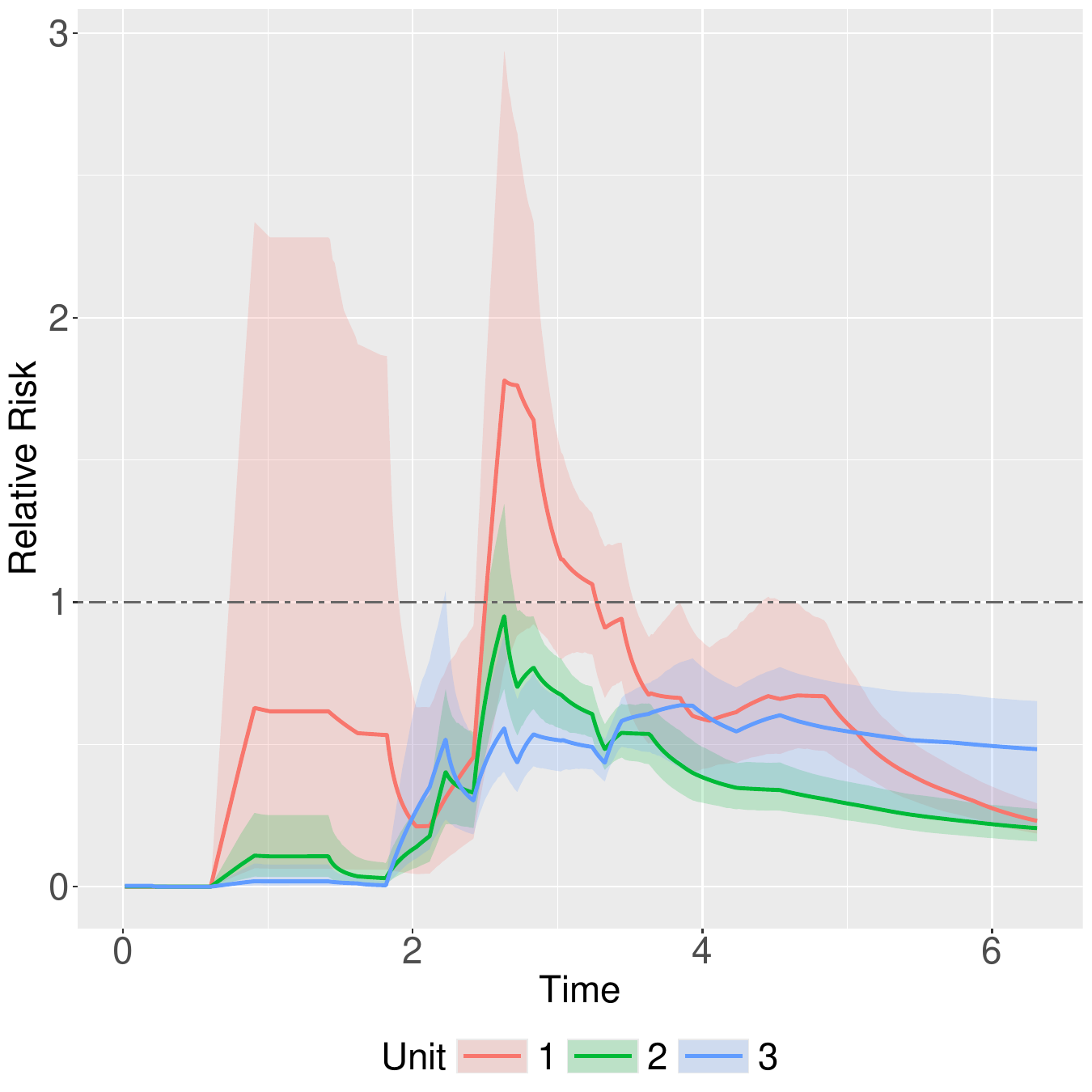} 
	\\
		(a) Overall cdf& (b) Relative risk\\
	\end{tabular}
	\caption{The predicted overall cdf and relative risk for three different GPU units.} \label{fig:real4}
\end{figure}

Other than the CIFs, the overall cdf can be used to predict the failure times. Additionally, conditioning on a failure time $t$, the probability $\Pr(C = k \mid T < t)$  for $k = 1, 2$ can be used to understand the relative risk among multiple failure types. Figure~\ref{fig:real4} (a) shows the overall cdf for three different GPU units in Fold 5. In terms of uncertainty quantification, Monte-Carlo dropout (\citeNP{pmlr-v48-gal16}) with dropout rate 0.4 and number of samples 1,000 is used to calculate the pointwise 95\% confidence intervals and the mean of the predictions, shown as the colored areas and colored lines in the figure. The covariates for Unit 1, Unit 2, and Unit 3 are cage 0, slot 7, node 2, column 15, row 8; cage 1, slot 8, node 3, column 11, row 1; and  cage 2, slot 1, node 2, column 3, row 4, respectively. The predicted overall cdf is the smallest for Unit 1, and is the largest for Unit 2.  Figure~\ref{fig:real4} (b) shows the relative risk of OTB versus DBE, which is calculated as   $\Pr(C = 1| T < t)/  \Pr(C = 2| T < t)$. The results show that the risk for OTB is higher when the time $t$ is approximately between $2.5$ years and $3.5$ years, for all three units. 

\section{Conclusions and Areas for Future Research}\label{sec:conclusion}

The failure time distribution prediction of complex systems with competing risks can be a challenging task.  In this study, we propose a deep learning model named SSH-Net to predict failure time distribution functions, including cause-specific hazards and CIFs, with competing risks. The proposed model associates the neural network structure with data structures, providing better interpretability of the network and enabling more efficient hyperparameter tuning informed by the data. Additionally, SSH-Net allows covariates from different system hierarchical levels to influence the underlying failure process separately through dedicated sub-networks. Furthermore, a smooth penalty term is incorporated into the loss function to prevent wiggly behavior in the cause-specific hazard function outputs. We demonstrate that our model outperforms alternatives in most scenarios using both simulation studies and the real-world Titan GPU dataset.

There are several contributions in this work. While some neural network models exist for predicting failure times with competing risks, they primarily focus on medical and clinical studies. For complex engineering systems, existing networks lack the flexibility to let different modules or hierarchies impact the failure time separately. Our model addresses this by processing different input groups through sub-networks before their effects are passed to a shared layer, allowing covariate groups to impact the failure separately. Secondly, hyperparameter tuning for neural networks is often time-consuming, especially with large datasets. Since we use piecewise constant hazard functions as outputs, the hyperparameters, bin number $J_k$ and smooth penalty $\lambda_k$ for the failure mode $k$, are associated directly with the duration of constant instantaneous failure rate in a period. Therefore, in real-world applications, researchers can pre-determine reasonable ranges of  $J_k$ and $\lambda_k$ based on domain knowledge to reduce the computational cost and to improve the prediction accuracy. Thirdly, existing studies rarely evaluate model performance with simulations with known underlying hazard functions. By designing a simulation that mimics the Titan GPU data, each model's performance against the true distribution functions are compared. From the simulation study, the proposed SSH-Net outperforms the two candidate models in terms of RMSE, Brier score, and AUC.

There are several future research directions. In this study, two connected layers are used to model the spatial location effect inside a supercomputer and their interactions. It would also be interesting to adopt the Gaussian process based neural network structure to capture the spatial dependencies among GPUs at different spatial locations for potential prediction improvement. Besides, as the sensor technology advanced, it is common to have time-varying sensory variables for complex systems. It would be beneficial to extend our work to allow time varying covariate inputs in the neural network model.

\section*{Acknowledgments}

The authors acknowledge Dr. Yili Hong for his valuable suggestions that helped improve the paper. The authors also thank the University of South Florida Research Computing for providing computational resources.

\section*{Appendix}
\begin{appendix}


	\section{Latent Failure Time Model and Simulation Setting} \label{apd1}
	To generate datasets that are similar to the Titan GPU data without favoring cause-specific model, we utilize the latent failure time model proposed in \shortciteN{Min03072025}. For a unit $j$ at cabinet location $l$, the latent failure time model assumes $T_{lj} = \min(T_{lj1}, T_{lj2})$, where $T_{ljk}$ is the latent failure time from failure type $k$. For notation simplicity, let $i$ denote the unit with location index $l$ and inside location index $j$.
	
	Following \shortciteN{Min03072025}, the latent failure times in the Titan GPU data are assumed to follow Weibull distributions. In particular, we let
	\begin{align*}
		f_{1}^L(t_{i} |\bx_j, w_{l1})= &\frac{1}{\xi_{1} t_{i}}\phi_{\SEV}\left [\frac{\log (t_{i})-\mu_{l1}}{\xi_{1}}\right], \\
		f_{2}^L(t_{i} | \bx_j, w_{l2})= &\lambda \frac{1}{\xi_{21} t_{i}}\phi_{\SEV}\left [\frac{\log (t_{i})-\mu_{l21}}{\xi_{21}}\right] +  (1 - \lambda) \frac{1}{\xi_{22} t_{i}}\phi_{\SEV}\left [\frac{\log (t_{i})-\mu_{l22}}{\xi_{2}} \right], 
	\end{align*}
	and
	\begin{align*}
		\mu_{l1} = \mu_1 + \bx_{jL}'\bbeta_1+w_{l1}, \; 	\mu_{l21} = \mu_2 + \bx_{jL}'\bbeta_2+w_{l2}, \; \mu_{l22} = \mu_{l21} + \eta,
	\end{align*}
	where $l = 1,2,\ldots, n_l$, and $j = 1, 2, \ldots, n_j$. The $\bx_{jL}$ are levels of categorical variables cage, slot, and node, and is different from $\bx_i$ in the neural network model, which contains additional information of cabinet location. The $n_l$ is the number of locations, which is set to be 200 in simulation, and $n_j$ is the total number of units at location $l$. The $\phi_{\SEV}$ denotes the pdf of the standard smallest extreme value (SEV) distribution, $\mu_1$, $\mu_{21},$ and $\mu_{22}$ denote the location parameters in the SEV distribution, $\xi_1$, $\xi_{21},$ and $\xi_{22}$ denote the scale parameters in the SEV distribution, $\bbeta_1$ and $\bbeta_2$ denote the coefficients correspond to the two different failure types, $w_{l1}$ and $w_{l2}$ denote normally distributed random effects that are used to capture the spatial information from cabinet location, $\lambda$ denotes the mixture probability for Failure type 2, and $\eta$ denotes the difference between the two location parameters for Failure type 2. The use of the mixture of two Weibull distributions for Failure type 2 is further explained in \shortciteN{Min03072025}. Additionally, the random effects are assumed to follow a multivariate normal distribution. In particular, let  $\bw_l=(w_{l1}, w_{l2})'$, and $\bw=(\bw_i', \ldots, \bw_n')'$. It is assumed that $\bw \sim \MVN(\boldsymbol{0},\Sigma_{\bf}\otimes \Omega)$,
	\begin{equation*}
		\Sigma =  \begin{pmatrix}
			\sigma_{1}^2 & \rho_{12}\sigma_{1} \sigma_{2}\\
			\rho_{12}\sigma_{1} \sigma_{2} & \sigma_{2}^2
		\end{pmatrix}, \; \Omega = (\omega_{sl}) = \exp\left[ - \left(\frac{d_{sl_c}}{\nu_c}\right)  - \left(\frac{d_{sl_r}}{\nu_r}\right) \right],\label{eq:colref}
	\end{equation*}
	where $\sigma_1$ and $\sigma_2$ are standard deviations of the random effects for two failure types, $\rho_{12}$ is the correlation coefficient between the random effects for the two failure types, $d_{sl_c}$  and $d_{sl_r}$ are the column and row distances between location $s$ and location $l$, and $\nu_c>0$ and $\nu_r>0$ are the parameters in the exponential correlation function. The values of true parameters are selected based on the estimates in \shortciteN{Min03072025} and are shown in Table~\ref{tab:sim.true}.
	\begin{table}
		\begin{center}
			\caption{True parameter values used in the simulation.}\label{tab:sim.true}
			\begin{tabular}{cccccccccccc}
				\hline\hline			 $\mu_1$ &   $\xi_1$  &$\mu_2$&   $\xi_{21}$& $\xi_{22}$ & $\lambda$ & $\eta$ & $\sigma_1$ & $\sigma_2$ & $\rho_{12}$             & $\nu_c$ & $\nu_r$       \\
				2.10 &  0.19 &  1.66 & 0.14 & 1.50 & 0.60 & 9.00 & 0.13& 0.11& 0.50& 0.60& 0.50\\ \hline
				\multicolumn{12}{c}{$\bbeta_1$}\\ 
				\multicolumn{12}{c}{  $(0.67 ,\; 0.27 ,   \;    0.04 ,      \; 0.03 ,   \;    0.05   ,  \;   0.04  ,  \;    0.07  ,  \;    0.01 ,     \; -0.01   ,    \; -0.29  ,   \;    -0.30,\; -0.06)'$} \\ \hline
				\multicolumn{12}{c}{$\bbeta_2$}\\ 
				\multicolumn{12}{c}{  $ (0.57    , \;    0.23     , \;    0.04  , \;        0.08      , \;   0.07   , \;      0.09     , \;    0.06      , \;   0.06   , \;      0.03   , \;      -0.24      , \;   -0.26, \; -0.06)'$} \\\hline\hline
			\end{tabular}
		\end{center}
	\end{table}

	\section{Transforming from Latent Failure Time Model to Neural Network Outputs} \label{apd2}
	Let $F_{1}^L(t \mid \bx_{jL}, w_{l1}) $ and $F_{2}^L(t \mid  \bx_{jL}, w_{l2})$ be the cdf for the latent failure times $T_{lj1}$ and $T_{lj2}$.  Let $S^L(t_{i1}, t_{i2} \mid \bx_{jL},\bw_l) = \Pr(T_{i1}>t_{i1}, T_{i2}>t_{i2} \mid \bx_{jL}, \bw_l)$ be the joint survival function. According to Theorem 3.1 in \citeN{crowder2001classical}, the cause-specific hazard defined in Section 2.1 is associated with the joint survival function, which is,
	\begin{align*}
		h^L(t,k\mid \bx_{jL},\bw_l) &= -\frac{\partial \log S^L(t_{i1}, t_{i2}\mid \bx_{jL}, \bw_l) }{\partial t_{ik}} \large \mid_{t_{i1} = t, t_{i2} = t}\\
		&=  - \frac{1}{S^L(t_{i1}, t_{i2} \mid \bx_{jL}, \bw_l)}\frac{\partial  S^L(t_{i1}, t_{i2}\mid \bx_{jL},\bw_l)}{\partial t_{ik}} \large \mid_{t_{i1} = t, t_{i2} = t}, \;k = 1, 2,
	\end{align*}
	where $L$ denotes the latent failure time model, distinguishing it from the predicted cause-specific hazards using neural networks.
	
	Note that conditioning on the random effects $\bw$, the latent failure times are independent.  Therefore,
	\begin{align*}
		S^L(t_{i1}, t_{i2}\mid \bx_{jL},\bw_l) &= \left[1 - F_{1}^L(t_{i1} \mid \bx_{jL}, w_{l1})\right][1 - F_{2}^L(t_{i2} \mid \bx_{jL}, w_{l2})],\\
		\frac{\partial  S^L(t_{i1}, t_{i2} \mid \bx_{jL},\bw_l)}{\partial t_{i1}} & =  -f_{1}^L( t_{i1} \mid \bx_{jl}, w_{l1}) [1 - F_{2}^L(t_{i2} \mid \bx_{jL}, w_{l2})], \\
		\frac{\partial  S^L(t_{i1}, t_{i2}\mid \bx_{jL},\bw_l) }{\partial t_{i2}} & =  -f_{2}^L( t_{i2} \mid \bx_{jl}, w_{l2}) [1 - F_{1}^L(t_{i1} \mid \bx_{jL}, w_{l1})], 
	\end{align*}
	and 
	\begin{align}
		h^L(t, 1 \mid  \bx_{jL},\bw_l) &= \frac{f_{1}^L(t \mid\bx_{jL}, w_{l1})}{1 - F_{1}^L(t \mid \bx_{jL}, w_{l1})}, 
		h^L(t, 2 \mid \bx_{jL},\bw_l) = \frac{f_{2}^L(t \mid \bx_{jL}, w_{l2})}{1 - F_{2}^L(t \mid \bx_{jL}, w_{l2})} \label{form:h2}, \\
		f^L(t,1 \mid \bx_{jL},\bw_l) &= f_{1}^L(t \mid\bx_{jL}, w_{l1})[1 - F_{2}^L(t \mid \bx_{jL}, w_{l2})], \label{form:h3} \\
		f^L(t,2\mid \bx_{jL},\bw_l) &= f_{2}^L(t \mid\bx_{jL}, w_{l2})[1 - F_{1}^L(t \mid \bx_{jL}, w_{l1})].\label{form:h4}
	\end{align}
	
	The corresponding CIFs, $F^L(t,k \mid \bx_{jL},\bw_l)$, can be calculated by integrating \eqref{form:h3} and \eqref{form:h4} numerically. Note that $F^L(t,k \mid \bx_{jL},\bw_l)$ is different from cdfs of latent failure times.
	In simulation studies, $h^L(t, k \mid  \bx_{jL},\bw_l)$ and $F^L(t,k \mid \bx_{jL},\bw_l)$ are the true values calculated from the latent failure time model.

	For SSH-Net, the outputs are the cause-specific hazards, which can be directly compared to the true values calculated based on \eqref{form:h2}. We denote the predicted cause-specific hazard as $\hat h^S(t,k \mid \bx_i)$ in this section. The CIF can be calculated by integrating the cause-specific pdf, which is
	\begin{align}
		\hat F^S(t,k \mid \bx_i) & = \int_{0}^t \hat f^S(u,k \mid \bx_i)du, \label{form:sh} \\
		\hat f^S(t,k \mid\bx_i) &= \hat h^S(t,k \mid \bx_i) \sum_{k = 1}^K \exp [-\hat H^S(t, k \mid \bx_i)] \notag,
	\end{align}
	where the calculation of $\hat H^S(t,k \mid \bx_i)$ is derived in Section 2.2, and the integration in \eqref{form:sh} can be calculated numerically.
	
	For NFG, the outputs are $b(\bx_i) = \widehat \Pr(G_i = k \mid \bx_i )$ from the balancing networks and $m(t, \bx_i)$ from the sub-distribution networks. The CIF from NFG is calculated as 
	\begin{align*}
		\hat F^N(t,k \mid \bx_i) = b(\bx_i) \{  1-  \exp[-t \times m(t,\bx_i)] \},
	\end{align*}
	which can be directly compared to $F^L(t,k \mid \bx_{jL},\bw_l)$. Let $q(t, \bx_i) =  -t \times m(t,\bx_i)$, the cause-specific hazard can be calculated as
	\begin{align}
		\hat h^N(t,k \mid \bx_i) = \frac{b(\bx_i) \exp[q(t, \bx_i)]}{[1 - \sum_{k = 1}^K \hat	F^N(t,k \mid \bx_i)]}  \frac{\partial q(t, \bx_i)}{\partial t} . \label{form:5}
	\end{align}
	In simulation studies, the derivative in \eqref{form:5} is calculated using numerical differentiation.
	
	For DeepHit, the outputs are the cause-specific pmfs under the discrete failure time assumption. Let $$(0, \delta, 2\delta, \ldots, (r-1)\delta)$$ be the discretized time grid, where $\delta$ is the step size, $r$ is the length of the time grid, and $(r-1)\delta = \tau$, the CIF is therefore calculated as 
	\begin{align*}
		\hat F^D(t, k \mid \bx_i) = \sum_{j = 0}^{r-1} \mathbbm{1}\left\{j\delta \leq t \right\} \widehat \Pr(t = j \delta, G_i = k \mid \bx_i),
	\end{align*}
	and can be compared with $F^L(t, k \mid \bx_i)$. A continuous analogue of the discrete hazard is calculated as
	\begin{align*}
		\hat h^D(j\delta, k \mid \bx_i) = \frac{1}{\delta}\frac{ \widehat \Pr(t = j \delta, G_i = k \mid \bx_i)}{1 - \sum_{k = 1}^K \hat F_i^D(t, k \mid \bx_i)},
	\end{align*}
	and is compared with $h^L(t, k \mid \bx_i)$. In the simulation studies, we let $\delta = 1/365$, as the original unit of the GPU failure time is year.
	
\section{Additional Simulation Results}
\begin{table}[]
	\centering
	\caption{Brier and Concordance score averaged on the time grid and across units in the test set. The scores are reported in units of 1e-2. }\label{aptab:brier}
	\begin{tabular}{cc|cccc}
		\hline\hline
		\multicolumn{2}{c|}{Shut Down Date}  & 2020-01-01 &  2019-01-01 &  2018-01-01 & 2017-06-01 \\
		\hline
		\multicolumn{6}{c}{$n$ = 10,000}\\
		\hline
		Failure Type 1 &	SSH-Net  & 4.65   &3.20  & 1.87  &   0.77\\
		Avg Brier (\%)	& NFG  &4.78    & 3.28 & 1.91 & 0.79  \\ \hline
		
		Failure Type 2 &SSH-Net  & 9.71   &7.07 & 4.38  &   1.85\\
		Avg Brier (\%)	& NFG  &9.90    & 7.19 & 4.46 & 1.88  \\ \hline
		Failure Type 1 &SSH-Net  & 82.17 & 82.39 & 81.46 &   82.21 \\
		Avg Concordance (\%)	& NFG  &79.80   & 79.87 & 77.29 & 70.55 \\
		\hline
		Failure Type 2 &	SSH-Net  & 78.10& 78.42 & 77.07 &  74.12 \\
		Avg Concordance (\%)	& NFG  &77.48   & 78.18 & 76.38 & 73.07 \\
		\hline
		\multicolumn{6}{c}{$n$ = 5,000}\\
		\hline
		Failure Type 1 &SSH-Net  & 5.60 & 4.19  & 2.32  &   1.10\\
		Avg Brier (\%)	& NFG  & 5.82   &4.32& 2.40 & 1.15 \\ \hline
		Failure Type 2 &SSH-Net & 10.36 & 7.57 & 4.62  &   2.09\\
		Avg Brier (\%)	& NFG  & 10.76   & 7.87& 4.79 & 2.17 \\ \hline
		Failure Type 1&	SSH-Net  & 80.19  & 81.29 &81.11 &   80.77\\
		Avg Concordance (\%)	& NFG  & 75.87   & 77.18 & 73.91& 66.67\\ \hline
		Failure Type 2&	SSH-Net  & 77.31  & 78.50 & 79.141 &   74.72\\
		Avg Concordance (\%)	& NFG  & 75.51   & 77.11 & 77.20& 71.03\\
		\hline\hline
	\end{tabular}
\end{table}
Table~\ref{aptab:brier} shows the comparison of the weighted Brier score and AUC between SSH-Net and NFG for all the simulation scenarios. The SSH-Net has smaller Brier and larger AUC, indicating that it outperforms NFG across all scenarios.

\end{appendix}	
\clearpage
\bibliographystyle{chicago}
\bibliography{ref}       

\end{document}